\documentclass[lettersize,journal]{IEEEtran}
\usepackage{hyperref}
\usepackage{orcidlink}
\usepackage{amsmath,amsfonts,amssymb}
\usepackage{multirow}
\usepackage[table]{xcolor} 
\usepackage{algorithmic}
\usepackage{algorithm}
\usepackage{array}
\usepackage{xcolor}
\usepackage[caption=false,font=normalsize,labelfont=sf,textfont=sf]{subfig}
\usepackage{textcomp}
\usepackage{stfloats}
\usepackage{url}
\usepackage{verbatim}
\usepackage{graphicx}
\usepackage{cite}
\usepackage{booktabs}
\usepackage{bm}
\newcommand{\eg}{\textit{e.g.}}
\newcommand{\ie}{\textit{i.e.}}

\begin{document}

\title{Spark3R: Asymmetric Token Reduction Makes Fast Feed-Forward 3D Reconstruction}

\author{Zecheng~Tang\textsuperscript{*},
        Jiaye~Fu\textsuperscript{*},
        Qiankun~Gao,
        Haijie~Li,
        Yanmin~Wu,
        Jiaqi~Zhang,
        Siwei~Ma,~\IEEEmembership{Fellow,~IEEE},
        and Jian~Zhang\textsuperscript{\dag}~\orcidlink{0000-0001-5486-3125},~\IEEEmembership{Member,~IEEE}
        \thanks{\textsuperscript{*}Zecheng Tang and Jiaye Fu contributed equally to this work.}
        \thanks{\textsuperscript{\dag}Corresponding author: Jian Zhang (e-mail: zhangjian.sz@pku.edu.cn).}
        \thanks{Zecheng Tang, Jiaye Fu, Qiankun Gao, Haijie Li, Yanmin Wu, and Jian Zhang are with the School of Electronic and Computer Engineering, Peking University, Shenzhen 518055, China.}
        \thanks{Jiaqi Zhang and Siwei Ma are with the School of Computer Science, Peking University, Beijing 100871, China.}
        }


\maketitle

\begin{abstract}
Feed-forward 3D reconstruction models based on Vision Transformers can directly estimate scene geometry and camera poses from a small set of input images, but scaling them to video inputs with hundreds or thousands of frames remains challenging due to the quadratic cost of global attention layers. 
Recent token-merging methods accelerate these models by
compressing the token sequence within the global attention
layers, but they apply a uniform reduction to query tokens
and key-value tokens, ignoring their functionally
distinct roles in 3D reconstruction. In this work, we identify a key property of feed-forward 3D reconstruction models: query tokens encode view-specific geometric requests and are sensitive to compression, while key-value tokens represent shared scene context and tolerate aggressive compression. Guided by this insight, we propose Spark3R, a training-free acceleration framework that decouples the compression of query tokens and key-value tokens by assigning distinct reduction factors, with intra-group token merging applied to query tokens and lightweight token pruning to key-value tokens. 
Additionally, Spark3R adaptively adjusts the key-value
reduction factor across layers, further improving the
quality--efficiency trade-off. As a plug-and-play framework requiring no retraining, Spark3R integrates directly into multiple pretrained feed-forward 3D reconstruction models, including VGGT, $\bm{\pi^3}$, Depth-Anything-3, and VGGT-$\bm{\Omega}$, and achieves up to $\mathbf{28\times}$ speedup on 1,000-frame inputs while maintaining competitive reconstruction quality.
\end{abstract}


\section{Introduction}
\label{sec:intro}

Recovering 3D geometry from multi-view images has traditionally relied on per-scene optimization pipelines such as Structure-from-Motion (SfM)~\cite{sfm, photo} and Multi-View Stereo (MVS)~\cite{mvs, mvsnet}. 
More recent neural scene representations, including NeRF~\cite{nerf} and 3D Gaussian Splatting~\cite{3dgs}, retain this per-scene paradigm and still demand substantial computational overhead.
Unlike per-scene optimization approaches~\cite{lu2024scaffold,gao2024hicom,fu2025recon}, feed-forward 3D reconstruction recovers dense scene geometry and camera poses directly from a set of unposed images in a single forward pass, with broad applications in autonomous driving~\cite{virpnet}, robotics~\cite{yang2026robo3r}, and multimodal 3D scene understanding~\cite{lmm_3d_understanding, 3ur_llm}. 
State-of-the-art models such as VGGT~\cite{vggt}, $\pi^3$~\cite{pi3}, Depth-Anything-3~\cite{depthanything3}, and VGGT-$\Omega$~\cite{vggt-omega} achieve this by processing all input frames jointly through global attention layers. 
While this joint processing yields accurate geometric consistency across an arbitrary number of frames, the global attention layers impose costs quadratic in the frame count, making it prohibitively expensive to scale to video-length inputs of hundreds or thousands of frames.

Streaming methods~\cite{cut3r,ttt3r,infinitevggt}
sidestep this cost through sequential processing but forgo
joint cross-frame reasoning, often degrading reconstruction
quality. Chunk-and-align
strategies~\cite{vggt-long,laser} partition the input into
fixed-size chunks, confining the model's own cross-frame
reasoning to within each chunk and relying on post-hoc
alignment of overlapping frames to recover global
consistency. Both lines of work gain
efficiency by restricting which frames can interact, rather
than accelerating the global attention layers themselves.


Rather than restricting which frames can interact,
token-merging methods~\cite{fastvggt,litevggt} offer a more
direct strategy: compressing the sequence length within
global attention layers so that all frames remain visible to
one another at reduced cost. However, these methods apply a
single, uniform reduction factor to all token roles, limiting speedups to roughly $10\times$ at best before
reconstruction quality degrades noticeably. We hypothesize
that this ceiling stems from ignoring the functionally
distinct roles of query tokens and key-value tokens in global
attention layers.

\begin{figure}[t]
  \centering
  \vspace{-6pt}
  \includegraphics[width=\linewidth]{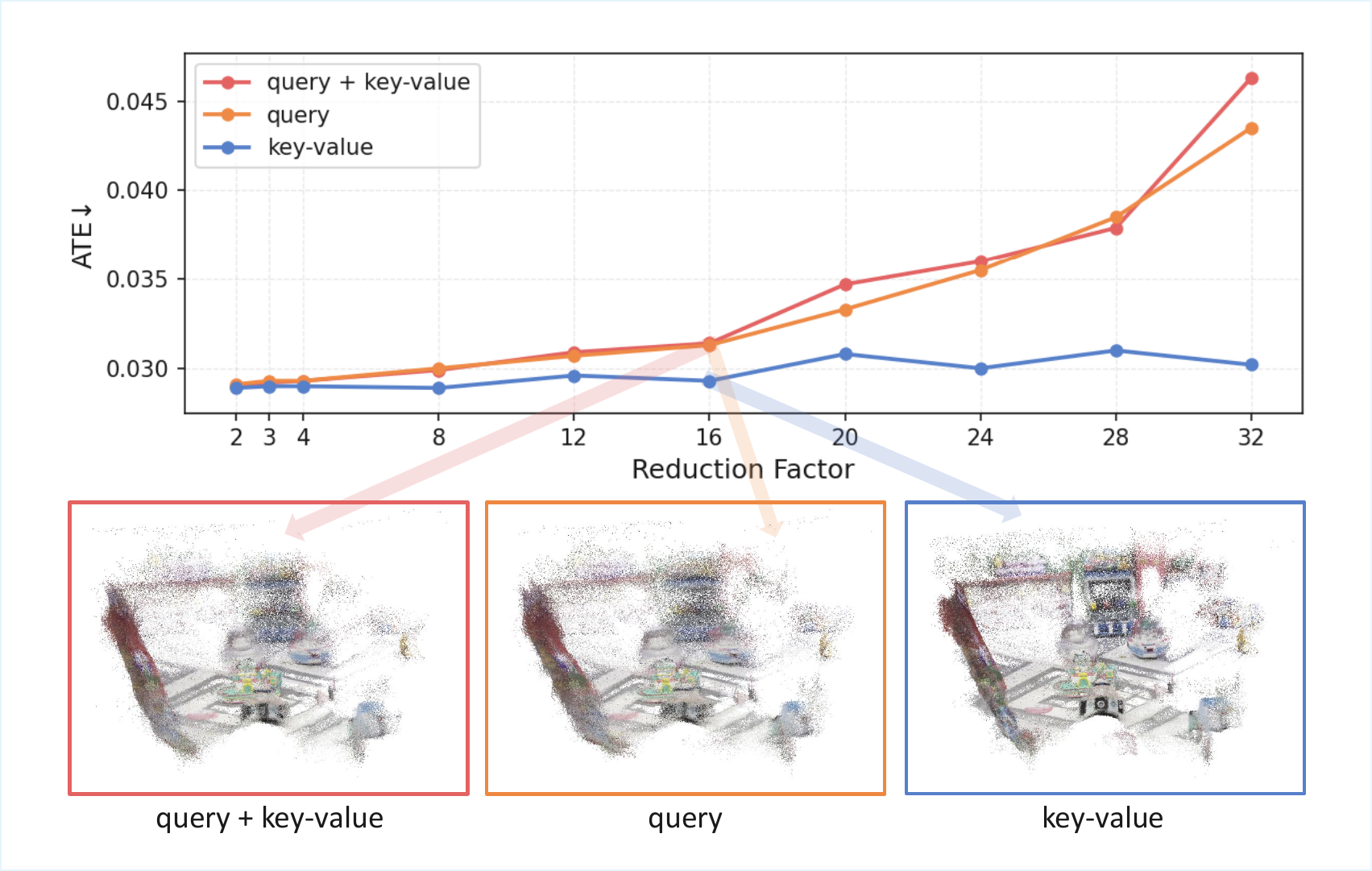}
  \vspace{-18pt}
  \caption{\textbf{Compression sensitivity of different token
  roles in VGGT.} We separately compress query tokens
  (orange), key-value tokens (blue), and both jointly (red)
  at increasing reduction factors and report pose error
  (ATE\,$\downarrow$). Key-value tokens tolerate aggressive
  compression with negligible quality loss, while query
  tokens degrade sharply beyond a reduction factor of~12.
  Joint uniform compression yields the steepest
  curve. \textit{Bottom}: point-cloud reconstructions at reduction factor~16.}
  \label{fig:qkv_sensitivity}
  
\end{figure}

To test this hypothesis, we separately compress query
tokens, key-value tokens, and both jointly at increasing
reduction factors (the ratio of the original sequence length to the reduced sequence length) and measure the resulting pose error on
VGGT. As shown in Figure~\ref{fig:qkv_sensitivity},
compressing key-value tokens alone keeps the error relatively
flat, whereas compressing
query tokens alone causes a sharp rise. Joint uniform
compression, the strategy adopted by existing token-merging methods,
yields the steepest curve of all: it inherits the full
sensitivity of query tokens while failing to exploit the
high compressibility of the key-value tokens. This pronounced asymmetry reveals substantial untapped
headroom that uniform methods leave on the table. We
provide complementary analytical explanations for this gap
in Sec.~\ref{sec:asymmetric}.

Guided by this insight, we present \textbf{Spark3R}, a
training-free, plug-and-play acceleration framework that
decouples the compression of query tokens and key-value tokens.
Our main contributions are:

\begin{itemize}
    \item We propose \textbf{asymmetric token reduction} for
        feed-forward 3D reconstruction models, which assigns
        separate reduction factors and distinct compression
        operators to query tokens and key-value tokens:
        intra-group token merging for query tokens and lightweight token pruning for key-value tokens.

    \item We introduce a \textbf{layer-adaptive key-value
        reduction schedule} that assigns heavier compression
        to empirically insensitive layers, further improving
        the quality--efficiency trade-off.

  \item We demonstrate that the resulting framework,
        \textbf{Spark3R}, integrates directly into VGGT,
        $\pi^3$, Depth-Anything-3, and VGGT-$\Omega$ \textbf{without
        retraining}, achieving up to $\mathbf{28\times}$
        speedup on 1{,}000-frame inputs while maintaining
        competitive reconstruction quality.
\end{itemize}
\section{Related Work}
\subsection{Feed-Forward 3D Reconstruction}
Feed-forward 3D reconstruction has emerged as a scalable
alternative to per-scene optimization methods such as
NeRF~\cite{nerf} and 3D Gaussian
Splatting~\cite{3dgs}. Early approaches such as
DUSt3R~\cite{dust3r} and MASt3R~\cite{mast3r} predicted
pairwise 3D point maps and fused them through post-hoc
global alignment.
VGGT~\cite{vggt} introduced an Alternating Attention
backbone that interleaves global cross-view self-attention
with frame-wise self-attention, enabling joint prediction of
depth, poses, and point clouds in a single forward pass.
$\pi^3$~\cite{pi3} proposed a fully
permutation-equivariant architecture that eliminates
dependence on a fixed reference view, and
Depth-Anything-3~\cite{depthanything3} showed that a plain
Vision Transformer backbone paired with a unified depth-ray
prediction target suffices for state-of-the-art geometry
reconstruction. 
VGGT-$\Omega$~\cite{vggt-omega} further scaled this paradigm to
billions of parameters and millions of sequences, achieving
state-of-the-art results on both static and dynamic benchmarks. 
To make such scaling tractable, it replaces a fraction of the
global attention layers with \emph{register attention}, in which
cross-frame interaction happens only among a small set of per-frame register tokens rather than among all image tokens.
Despite these architectural advances, all of the above
models still rely on attention layers whose cost grows
quadratically with the total token count, creating a
fundamental scalability bottleneck as the number of input
frames grows.

\subsection{Streaming 3D Reconstruction}
Streaming methods address the quadratic scaling cost of global attention layers by processing
frames incrementally. Recurrent-state approaches such as
CUT3R~\cite{cut3r} and TTT3R~\cite{ttt3r} maintain a
compact state updated via cross-attention with each incoming
frame, yielding constant per-step cost but progressively
losing early context. 
Explicit-memory methods such as
Point3R~\cite{point3r} and
StreamVGGT~\cite{streamvggt} anchor memory at reconstructed
3D positions or cache historical keys and values, trading
linear memory growth for richer context retention.
Chunk-and-align methods~\cite{vggt-long,laser} partition
the input into fixed-size windows, run inference on each chunk separately, and stitch the resulting submaps via post-hoc
$\mathrm{Sim}(3)$ registration. All these approaches sacrifice cross-frame reasoning scope
for speed.

\subsection{Efficient Feed-Forward 3D Reconstruction}

A complementary line of work preserves the global-context paradigm but reduces per-forward-pass cost.
Two principal strategies have been explored.

Token-merging methods exploit the observation that global
attention maps exhibit substantial redundancy, with many
tokens attending to nearly identical patterns, and merge
similar tokens before the attention computation.
FastVGGT~\cite{fastvggt} is the first to apply token merging
to VGGT, achieving roughly $4{\times}$ speedup with no
retraining by merging similar tokens via cosine similarity.
LiteVGGT~\cite{litevggt} introduces geometry-aware
partitioning and further accelerates VGGT by reusing
merge indices across adjacent layers and incorporating fp8
fine-tuning, reaching approximately $10{\times}$ speedup.
Both methods apply a single, uniform reduction factor to all
tokens, overlooking the distinct roles of
query tokens versus key-value tokens.

TTT-based methods take a different approach by replacing
quadratic softmax attention with linear-time operations.
{VGG-T$^{3}$}~\cite{vggt3} substitutes each global attention
layer with a test-time training (TTT) module, scaling linearly
with the number of frames.
ZipMap~\cite{zipmap} extends this with large-chunk TTT
layers~\cite{lact} interleaved,
achieving over $20{\times}$ speedup on 750-frame inputs.
Both methods substantially outperform streaming approaches
in reconstruction quality but still trail
$\pi^{3}$~\cite{pi3} on most benchmarks, and incur
considerable training overhead: VGG-T$^{3}$ requires 100k
steps on 8~A100s and ZipMap requires 180k steps on 64~H100s.

Spark3R shares the token-compression philosophy of
token-merging methods but introduces asymmetric
token reduction and a layer-adaptive key-value reduction schedule,
achieving speedups comparable to ZipMap's while requiring no
retraining---unlike TTT-based methods that incur substantial
training cost to replace the attention mechanism entirely.

\section{Preliminaries}

\subsection{Scaled Dot-Product Attention}
\label{sec:sdpa}

Scaled dot-product attention (SDPA)~\cite{attn} is the core
building block of Transformer architectures. Given an input
token sequence
$\mathbf{X} \in \mathbb{R}^{N \times D}$, three linear
projections produce the query, key, and value matrices
$\mathbf{Q}$, $\mathbf{K}$, $\mathbf{V}
\in \mathbb{R}^{N \times d}$. The attention output is:
\begin{equation}
    \operatorname{Attention}(\mathbf{Q}, \mathbf{K}, \mathbf{V})
    = \operatorname{softmax}\!\left(\frac{\mathbf{Q}\mathbf{K}^{\top}}{\sqrt{d}}\right)\mathbf{V}.
    \label{eq:sdpa}
\end{equation}
The time and memory complexity of SDPA are both $\mathcal{O}(N^2)$; although FlashAttention~\cite{flash-attn} reduces the memory complexity to linear in $N$, the quadratic time complexity remains.

\subsection{Feed-Forward 3D Reconstruction Models}
\label{sec:feedforward}

Feed-forward 3D reconstruction models predict dense 3D
scene properties---depth maps, camera poses, and point
clouds---from unposed images in a single forward pass.
Representative works include VGGT~\cite{vggt},
$\pi^3$~\cite{pi3}, and
Depth-Anything-3~\cite{depthanything3}. Despite differences
in architecture and training, all three follow a common
three-stage pipeline, described below using VGGT as the
representative example.

\textbf{Image Encoder.}
A pretrained vision foundation model (\eg, DINOv2~\cite{dinov2})
independently extracts $L$ patch tokens per frame. With a
learnable camera token and four register tokens appended,
each frame has $P = L + 5$ tokens. Stacking all $S$ frames
produces
$\mathbf{F} \in \mathbb{R}^{B \times S \times P \times D}$.

\textbf{Alternating Attention Backbone.}
The backbone alternates two forms of self-attention. Frame
attention layers operate within each frame's $P$ tokens at
a cost of $\mathcal{O}(S \cdot P^2)$, linear in~$S$.
Global attention layers concatenate all tokens into a single
sequence of length $N = S \times P$ and perform full
pairwise interaction at a cost of
$\mathcal{O}(N^2) = \mathcal{O}(S^2 \cdot P^2)$, quadratic
in both frame count and per-frame token count. As the input
grows, this quadratic scaling dominates inference cost.

\textbf{Task-Specific Prediction Heads.}
Lightweight heads decode the backbone outputs into 3D
predictions: a camera head regresses per-frame intrinsics
and extrinsics from special tokens, and DPT-based
heads~\cite{dpthead} decode patch tokens into depth maps and
point clouds.

\subsection{Token Merging}
\label{sec:tome}

Token Merging (ToMe~\cite{tome}) is a training-free
technique that reduces the effective sequence length in
Vision Transformers by merging redundant tokens. Given an
input sequence
$\mathbf{X} = \{x_1, \dots, x_N\}$, the tokens are
partitioned into a destination set $\mathcal{D}$ and a
source set $\mathcal{S}$ using strategies such as
fixed-stride sampling in ToMe~\cite{tome} or region-based random
sampling in ToMeSD~\cite{tomesd}. Each source token
$x_s \in \mathcal{S}$ is then matched to its most similar
destination token $x_d \in \mathcal{D}$ via cosine similarity:
\begin{equation}
    \mathrm{sim}(x_s, x_d)
    = \frac{x_s \cdot x_d}{\|x_s\|\,\|x_d\|}.
\end{equation}
A destination token $x_d$ matched by $n$ source tokens is
updated as their average:
\begin{equation}
    x_d' = \frac{x_d + \sum_{i=1}^{n} x_{s_i}}{n + 1}.
    \label{eq:tome_avg}
\end{equation}
Source tokens are discarded, reducing the sequence length for
the subsequent attention layers. For dense prediction tasks,
an unmerging step restores the original length by copying
each destination's updated representation back to its
matched sources.
\section{Method}
\begin{figure}[t]
  \centering
  \vspace{-4pt}
  \includegraphics[width=\linewidth]{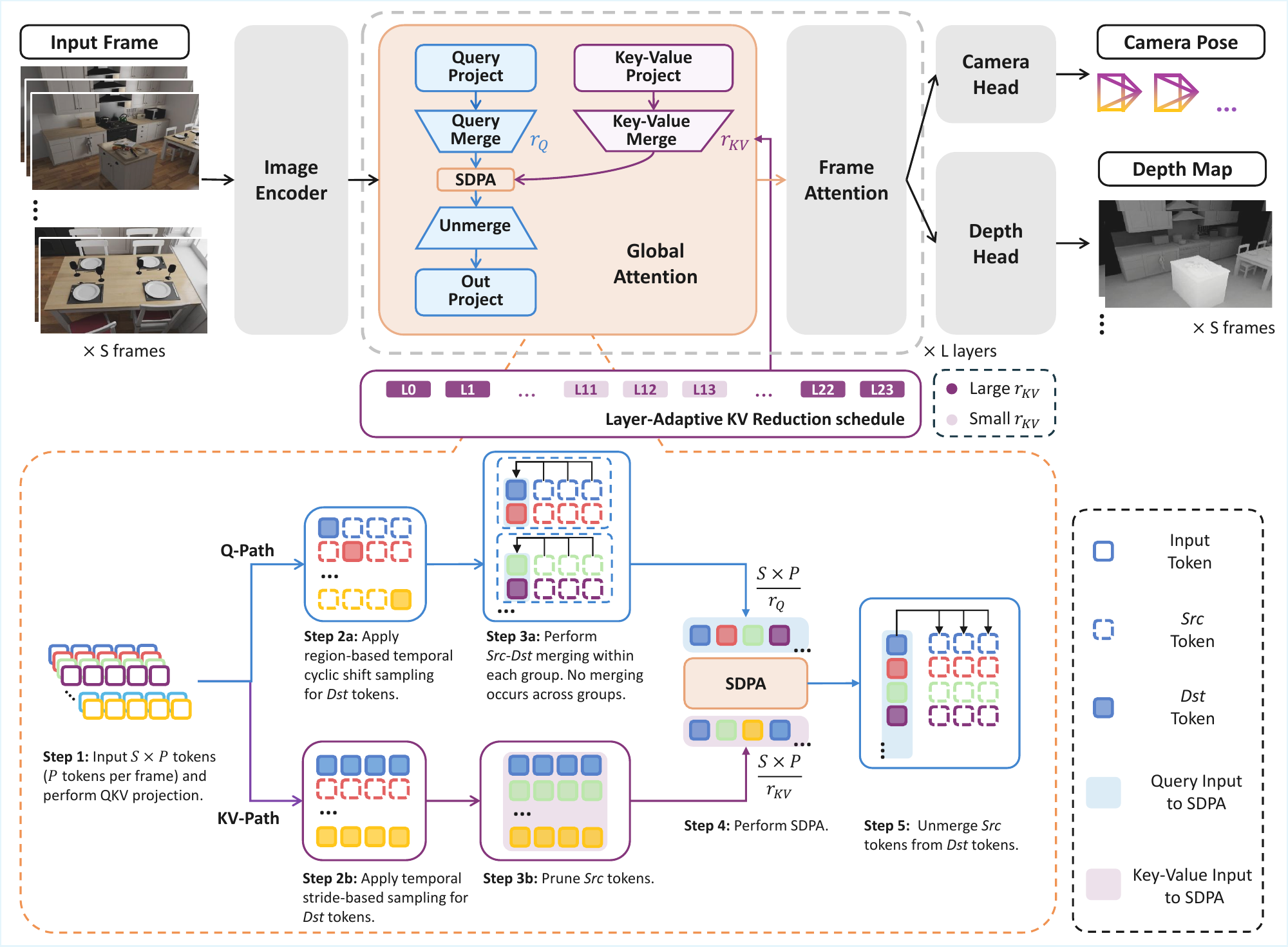}
  \vspace{-12pt}
  \caption{\textbf{Overview of Spark3R.}
(\textit{Top})~Spark3R applies \textbf{asymmetric token
reduction} to the global attention layers of a feed-forward
3D reconstruction model, with separate reduction factors
$r_{\text{Q}}$ and $r_{\text{KV}}$
($r_{\text{KV}} > r_{\text{Q}}$ in general).
(\textit{Middle})~A \textbf{layer-adaptive key-value reduction schedule} assigns
each layer a large or small $r_{\text{KV}}$ based on its
measured sensitivity to compression.
(\textit{Bottom})~Detailed illustration of the asymmetric
reduction. Tokens from different frames are distinguished by
color. The Q-Path performs intra-group token merging; the
KV-Path applies lightweight token pruning. Source tokens are
restored via unmerging after SDPA.}
  \label{fig:main}
\end{figure}

We propose Spark3R, a token-compression
framework with three complementary designs, as illustrated in
Figure~\ref{fig:main}.
First, \emph{asymmetric token reduction} assigns separate reduction
factors to query tokens and key-value tokens, reflecting their distinct
sensitivities to compression (Sec.~\ref{sec:asymmetric}).
Second, dedicated compression operators are tailored to each token type: \emph{intra-group token merging} for query tokens (Sec.~\ref{sec:qmerge}) and \emph{lightweight token pruning} for key-value tokens (Sec.~\ref{sec:kvmerge}).
Third, a \emph{layer-adaptive key-value reduction schedule} adjusts the key-value reduction factor of
each global attention layer according to its measured sensitivity to
compression (Sec.~\ref{sec:layerwise}).

\subsection{Asymmetric Token Reduction}
\label{sec:asymmetric}

The central insight of Spark3R is that query tokens and
key-value tokens in global attention layers exhibit
fundamentally different sensitivities to compression.
Sec.~\ref{sec:intro} and
Figure~\ref{fig:qkv_sensitivity} have already demonstrated
this asymmetry empirically; here we provide two
complementary analytical explanations that clarify
\emph{why} the gap exists and inform our design choices:
one grounded in the semantics of 3D reconstruction, the
other in the structure of the attention computation.

\textbf{Reconstruction Semantics.}
As discussed in Sec.~\ref{sec:feedforward}, each output
of a global attention layer must carry sufficient
information for the downstream prediction heads to produce
view-specific estimates (\eg, depth and camera pose) for
every input frame. Because query tokens dictate which
outputs are produced, each query token encodes a
viewpoint-specific geometric request: two query tokens that
observe the same 3D surface point from different cameras
must ultimately yield distinct depth values and camera
parameters. Compressing such tokens conflates requests that
are spatially proximate yet semantically distinct, directly
degrading per-view predictions. Key-value tokens, by
contrast, collectively form a shared representation of the
underlying 3D scene. When multiple frames observe the same
region, their key-value tokens encode largely overlapping
geometric information and are therefore naturally amenable
to aggressive compression.

\textbf{Attention Structure.}
The algebraic structure of SDPA (Eq.~\eqref{eq:sdpa})
reinforces this asymmetry. Because the attention output has
one row per query token, reducing the query sequence
directly shrinks the output and requires an unmerging step
to restore the original length, introducing approximation
error at every restored position. Reducing the key-value
sequence, by contrast, only coarsens the context each query
attends over, and can therefore be
applied at considerably higher reduction factors.

Together with the empirical evidence in
Figure~\ref{fig:qkv_sensitivity}, these observations
motivate our decoupled design: rather than applying a
single uniform reduction factor, we assign separate
factors $r_{\text{Q}}$ and $r_{\text{KV}}$
($r_{\text{KV}} > r_{\text{Q}}$ in general) to query
tokens and key-value tokens, respectively.

\subsection{Intra-Group Query Merging}
\label{sec:qmerge}

Having established that query tokens are considerably more
sensitive to compression than key-value tokens, we now
describe the query compression operator, which must
preserve reconstruction accuracy under a limited reduction
budget. We adopt the token merging framework
(Sec.~\ref{sec:tome}) as our starting point and
introduce \emph{intra-group matching} to reduce its
computational overhead while maintaining merge quality.

\begin{figure}[t]
\centering
\vspace{-2pt}
\includegraphics[width=\linewidth]{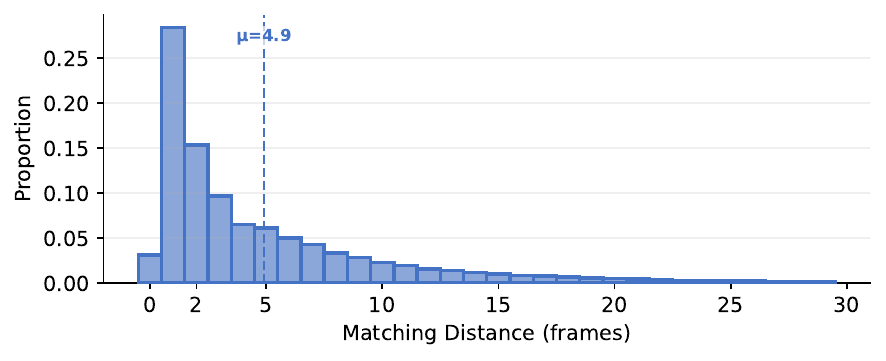}
\vspace{-12pt}
\caption{Distribution of inter-frame distances between
merged source--destination token pairs produced by standard token merging on
query tokens.}
\label{fig:merge_distance}
\end{figure}

\textbf{Locality of Merge Pairs.}
Standard token merging partitions the full concatenated
sequence of $N = S \times P$ tokens into source and
destination sets and computes similarities between them,
where $S$ denotes the number of frames and $P$ the number of tokens per frame. 
We observe, however, that the resulting merges are overwhelmingly \emph{local}. 
To quantify this, we measure the inter-frame distance of every merged
pair produced by applying standard token merging to query
tokens. As shown in Figure~\ref{fig:merge_distance},
merges within the same frame are rare, the distribution
peaks sharply at distance~1, and virtually no merges occur
beyond 20~frames. This strong locality indicates that the
full $\mathcal{O}(S^{2})$ global matching is unnecessary:
nearly all beneficial matches can be discovered within a
small temporal neighborhood.

\textbf{Intra-Group Matching.}
Motivated by this observation, we partition the $S$ input
frames into $\lceil S/G \rceil$ non-overlapping groups of
$G$ consecutive frames and perform token merging
independently within each group of $G \times P$ tokens.
This reduces the matching cost from
$\mathcal{O}(S^{2})$ to $\mathcal{O}(S \cdot G)$,
which scales linearly in~$S$ for a fixed group size~$G$.
By exploiting the locality identified above, intra-group
matching achieves comparable merge quality to the global
strategy while reducing the matching complexity from
quadratic to linear in the sequence length.

\textbf{Configuration.}
Within each group, destination tokens are selected by cycling through a fixed set of spatial offsets across consecutive frames, so that each frame contributes a different, deterministic subset of its tokens as destinations. This is a deterministic cyclic variant of the region-based random
sampling used in ToMeSD~\cite{tomesd} and FastVGGT~\cite{fastvggt}, yielding
comparable merge quality while being slightly more efficient to compute.
We set the group size to $G = 20$ as a practical
trade-off between matching quality and computational cost.
The query reduction factor is scaled with the input length:
\begin{equation}
    r_{\text{Q}} = \begin{cases}
        1 & \text{if } S \leq 100, \\
        2 & \text{if } 100 < S \leq 300, \\
        3 & \text{if } 300 < S \leq 500, \\
        4 & \text{if } S > 500.
    \end{cases}
    \label{eq:adaptive_rq}
\end{equation}
We ablate the group size choice in
Sec.~\ref{sec:ablation}.

\subsection{Lightweight Key-Value Pruning}
\label{sec:kvmerge}

We now turn to the compression of key-value tokens. Since
key-value tokens tolerate substantially higher reduction
factors than query tokens, we can simplify not only the
reduction factor but also the compression operator itself.

\begin{figure}[b]
  \centering
  \vspace{-3pt}
  \includegraphics[width=\linewidth]{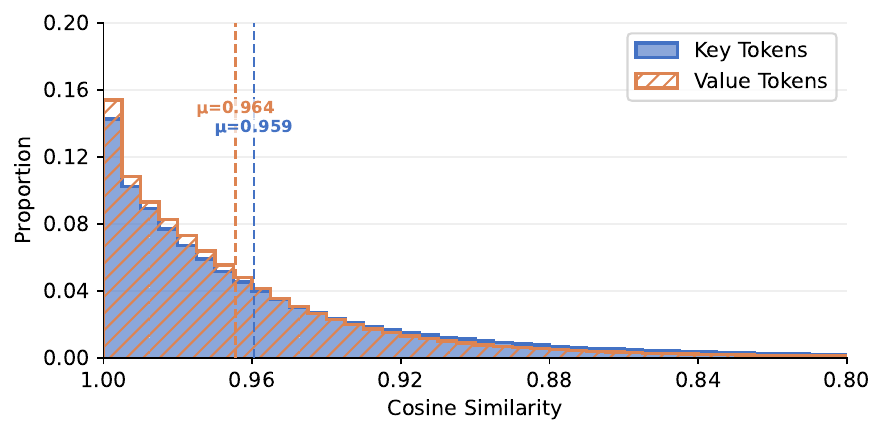}
  \vspace{-18pt}
  \caption{Distribution of cosine similarities between
  matched source--destination token pairs produced by
  standard token merging, measured separately for key tokens and value tokens.}
  \label{fig:kv_sim}
\end{figure}

\textbf{From merging to pruning.}
Recall that in standard token merging, each destination
token $x_d$ is updated as the average of itself and its $n$
matched source tokens.
Because each source token is matched to its most similar
destination token, the cosine similarity within matched
pairs is extremely high in practice.
Figure~\ref{fig:kv_sim} confirms this empirically: for both
key tokens and value tokens, the vast majority of matched
source--destination pairs exceed a cosine similarity
of~0.9, so the averaging only
marginally affects destination representations. The destination update can therefore be dropped: destination
tokens retain their original values while source tokens are
simply discarded.

Without the averaging update, which destination each source token was matched to is never used, so the similarity computation, which is the dominant cost of the matching step, 
can be skipped entirely, 
reducing the compression operator to lightweight token pruning with negligible overhead.

\textbf{Configuration.}
Because the compression reduces to pure pruning without
averaging or unmerging, the only remaining design choice is
which tokens to retain. In practice, this choice is fairly
forgiving: even random token-level sampling yields
competitive results. We find that temporal stride sampling,
which retains all tokens from every $r_{\text{KV}}$-th
frame and discards the rest, performs slightly better while
being deterministic and trivial to implement. 
We further scale $r_{\text{KV}}$ with the input length to
maintain a roughly linear computation budget:
\begin{equation}
  r_{\text{KV}} =
  \begin{cases}
    1 & \text{if } S \le 100,\\
    \lceil S / 40 \rceil & \text{otherwise},
  \end{cases}
  \label{eq:adaptive_rkv}
\end{equation}
where $S$ denotes the number of input frames.
Together, Eqs.~\eqref{eq:adaptive_rq} and~\eqref{eq:adaptive_rkv} form a length-adaptive reduction schedule that scales compression with the number of input frames, applying minimal or no reduction on short sequences while enabling aggressive compression on long ones.

\subsection{Layer-Adaptive Key-Value Reduction Schedule}
\label{sec:layerwise}

As noted in prior work on KV-cache
compression~\cite{fastkv}, different layers can exhibit vastly
different sensitivities to key-value compression. We observe
the same phenomenon in feed-forward 3D reconstruction
models: most layers tolerate aggressive key-value pruning
with negligible impact on accuracy, while a subset
suffers substantial degradation under the same reduction
factor. This motivates an adaptive strategy that assigns
different key-value reduction factors to different layers.

\begin{figure}[t]
  \centering
  \vspace{-2pt}
  \includegraphics[width=\linewidth]{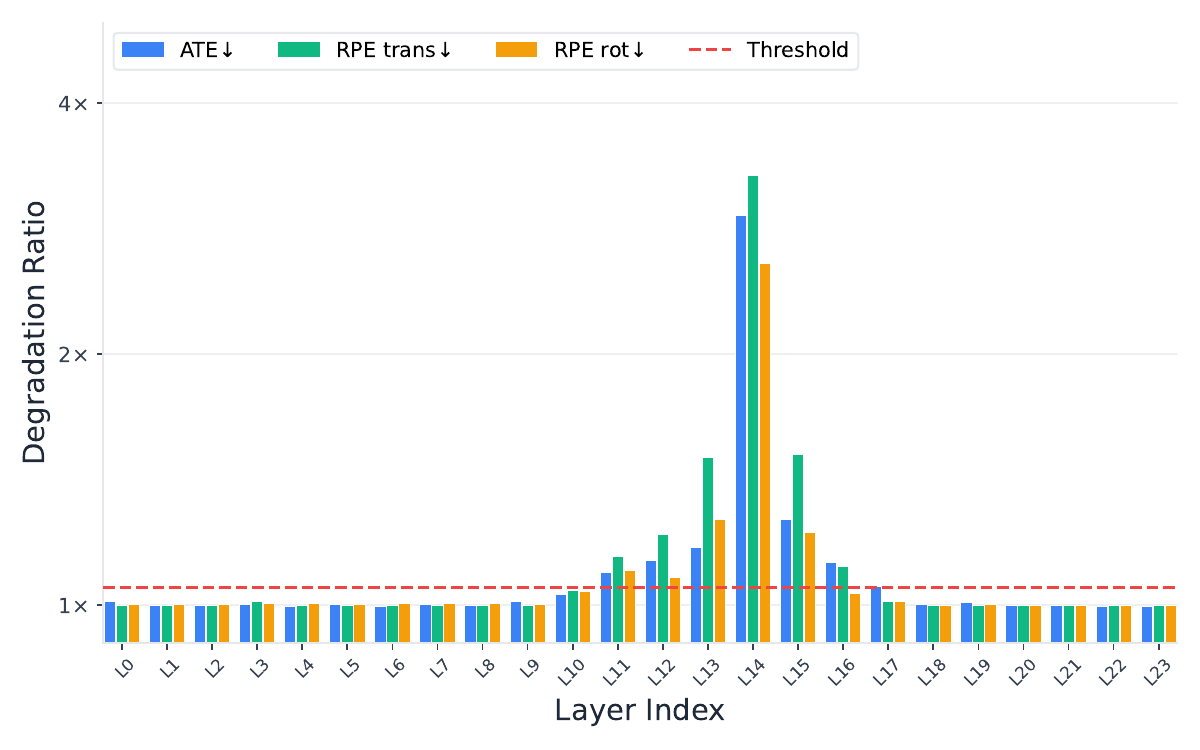}
  \vspace{-12pt}
  \caption{
    Per-layer sensitivity to key-value pruning in VGGT
    (24~layers). Each group of bars reports the degradation
    ratio when a single layer's reduction factor is raised
    from~$32$ to~$256$, with all other layers held at~$32$.
    The red dashed line marks the $1.05{\times}$~threshold
    used to separate high- and low-sensitivity layers.
  }
  \label{fig:vggt_layerwise}
\end{figure}

\begin{figure}[t]
  \centering
  \includegraphics[width=0.32\linewidth]{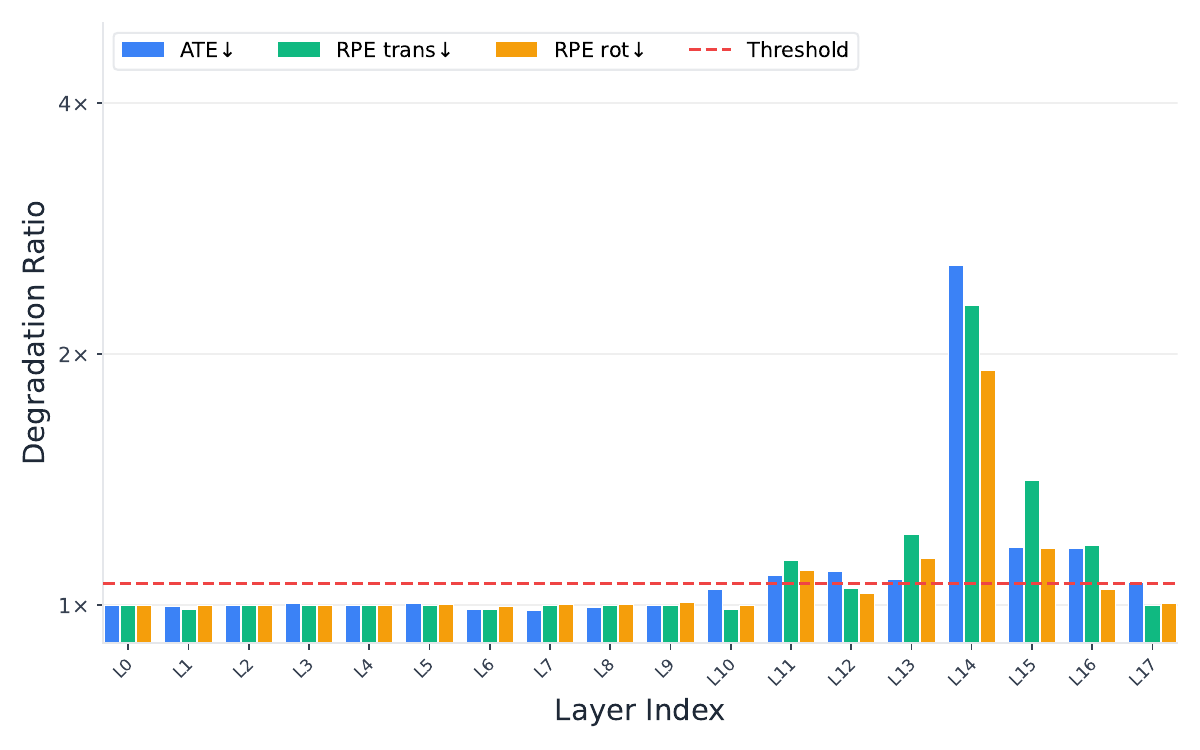}
  \hfill
  \includegraphics[width=0.32\linewidth]{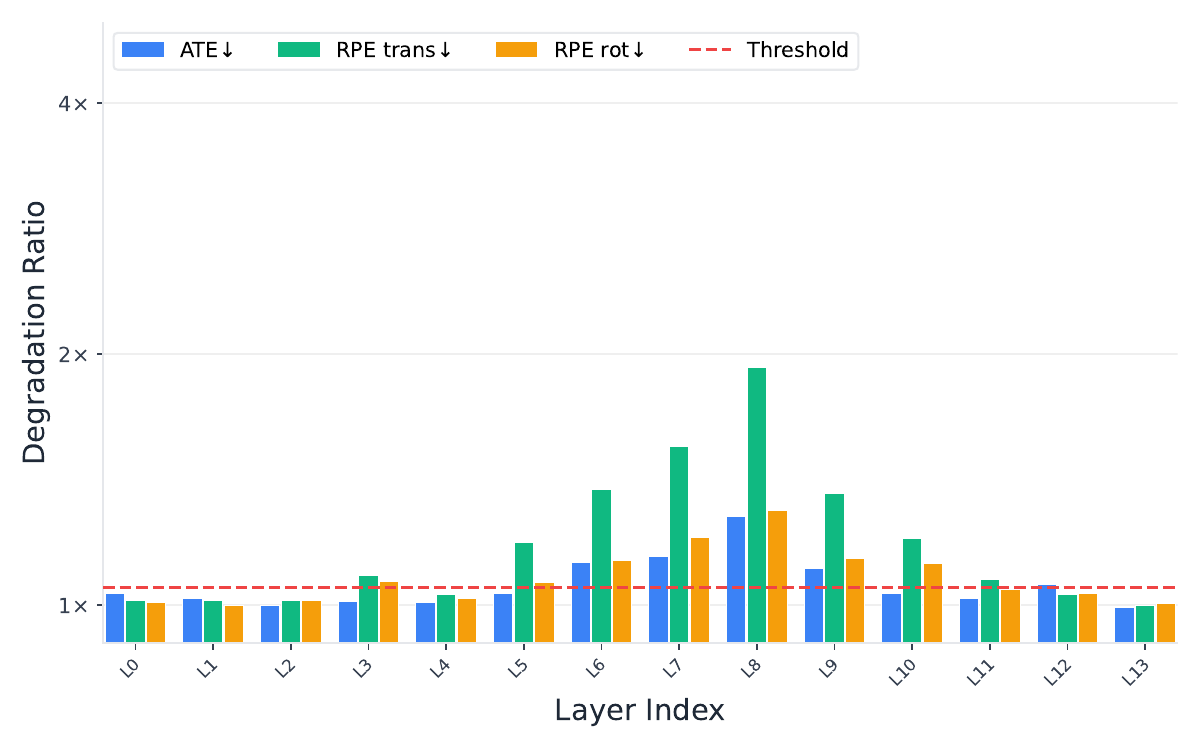}
  \includegraphics[width=0.32\linewidth]{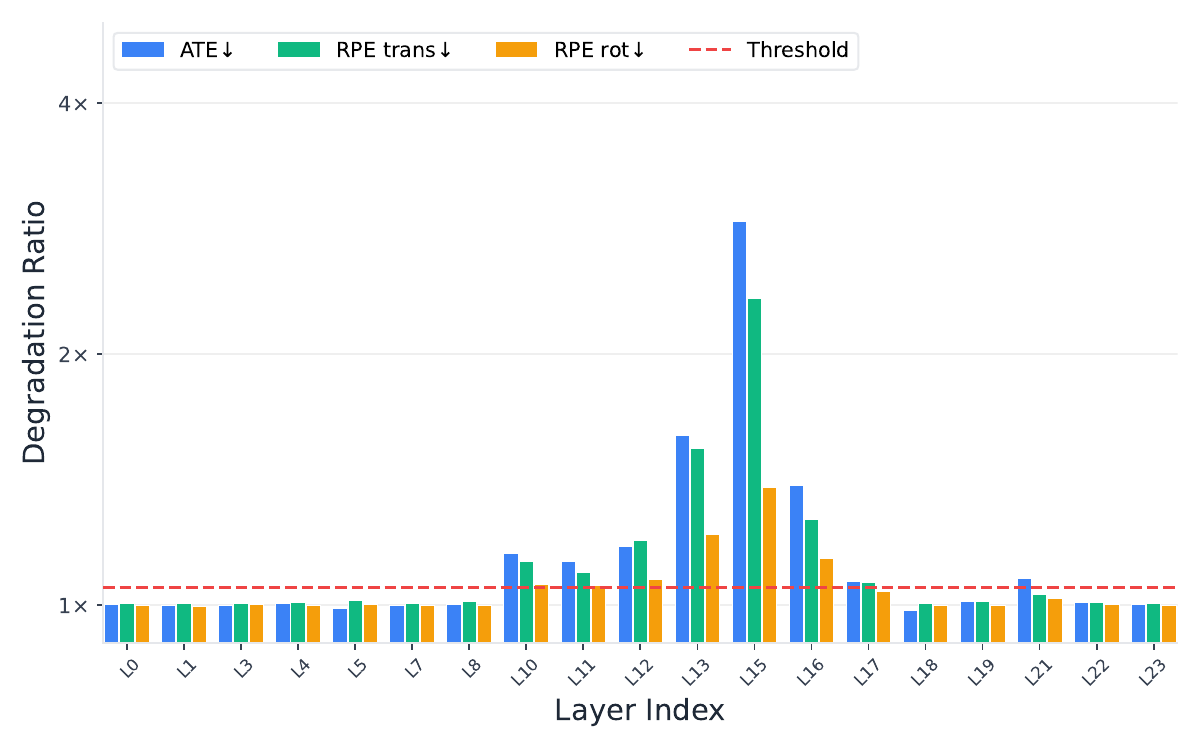}
  \caption{
      Per-layer sensitivity to key-value pruning in
      $\pi^3$ (\textit{left}, 18~layers),
      Depth-Anything-3 (\textit{middle}, 14~layers) and VGGT-$\Omega$ (\textit{right}, 24~layers).
      Each group of bars is obtained with the same protocol
      as Figure~\ref{fig:vggt_layerwise}.
      Zoom in for better view.
    }
  \label{fig:pi3_da3_layerwise}
\end{figure}

To identify which layers are sensitive, we conduct a
layer-wise probing experiment. Starting from a uniform
reduction factor of~$r_{\text{KV}}{=}32$ applied to all
layers, we raise the reduction factor of a single target
layer to~$256$ while keeping all other layers at~$32$, and
measure the resulting degradation in prediction accuracy.
This procedure is repeated independently for every global
attention layer (in VGGT-$\Omega$, this excludes 5
register attention layers at indices 2, 6, 9, 14, 20,
which attend only over per-frame register tokens);
the per-layer degradation ratios for VGGT are shown in
Figure~\ref{fig:vggt_layerwise}. Results for $\pi^3$,
Depth-Anything-3, and VGGT-$\Omega$ (Figure~\ref{fig:pi3_da3_layerwise}) exhibit similar patterns.

The results reveal highly non-uniform sensitivity across all
four models. Sensitive layers tend to concentrate in the
middle-to-late portion of the models: layers 11--16 in both
VGGT and~$\pi^3$, layers 3--12 in Depth-Anything-3, and layers 10--13, 15--17, and 21 in VGGT-$\Omega$. 
The remaining layers stay below a degradation ratio
of~$1.05{\times}$,
indicating that they can absorb substantially more
aggressive pruning without meaningful accuracy loss.

We therefore partition all layers into two tiers based on
whether their degradation ratio exceeds~$1.05{\times}$:
\begin{itemize}
  \item \textbf{High-sensitivity layers} retain the base
    key-value reduction factor~$r_{\text{KV}}$ to preserve
    prediction quality.
  \item \textbf{Low-sensitivity layers} are assigned an
    elevated reduction factor of~$l \cdot r_{\text{KV}}$,
    where the multiplier~$l$ controls how aggressively
    these layers are pruned. We set $l{=}3$ by default;
    ablations on this choice are provided in
    Sec.~\ref{sec:ablation}.
\end{itemize}


This simple two-tier, layer-granularity scheme requires only a single offline probing pass per base model and adds no runtime overhead. As shown in the experiments
(Sec.~\ref{sec:experiments}) and ablations
(Sec.~\ref{sec:ablation}), the resulting schedule
consistently improves the quality--efficiency trade-off
across all four base models and all evaluated tasks.

\section{Experiments}
\label{sec:experiments}

We integrate Spark3R into three representative feed-forward
3D reconstruction models, namely VGGT~\cite{vggt},
$\pi^3$~\cite{pi3}, and Depth-Anything-3
(DA3, Giant variant)~\cite{depthanything3}, and evaluate the resulting
quality--efficiency trade-offs across architectures. To
further evaluate Spark3R on newer backbones, we additionally
report results on VGGT-$\Omega$~\cite{vggt-omega}, a model
released after our initial submission; these results are
presented separately throughout to keep the main comparison
aligned with the submission-time baselines.
Qualitative comparisons are provided in
Sec.~\ref{sec:qual}, and ablation studies are presented in
Sec.~\ref{sec:ablation}.

\textbf{Evaluation Protocol.}
Extending the protocol of TTT3R~\cite{ttt3r},
we retain \emph{all} available frames up to a maximum of
1{,}000, so that the benchmark directly evaluates each
method's ability to reconstruct long sequences at scale.
For the point map benchmarks (7-Scenes, NRGBD) and long-sequence camera pose benchmarks (TUM-dynamics, ScanNet), this protocol yields sequences of approximately 1{,}000 frames, considerably longer than those used in prior evaluations and
creating a demanding testbed that stresses the 
scalability of each method.
The remaining benchmarks consist of shorter sequences:
Bonn (video depth) averages approximately 623 frames,
KITTI (video depth) approximately 264, and Sintel
(camera pose and video depth) only approximately 46.
This diversity also serves to verify that Spark3R's 
length-adaptive reduction schedule in
Eqs.~\eqref{eq:adaptive_rq}--\eqref{eq:adaptive_rkv}
introduces minimal quality degradation on shorter inputs.

\textbf{Baselines and Implementation Details.}
We compare against three categories of methods:
(i)~the unaccelerated feed-forward models themselves
(VGGT, $\pi^3$, and DA3), which serve as quality upper
bounds; (ii)~existing acceleration methods for feed-forward
models, including FastVGGT~\cite{fastvggt} and
ZipMap~\cite{zipmap}, the latter being the current
state-of-the-art approach that reports over $20{\times}$
speedup over VGGT at 750~frames with linear scaling, albeit
at the cost of fine-tuning a larger 1.4B model for 180K
steps on 64 H100 GPUs;
and (iii)~streaming 3D reconstruction models that process
frames sequentially, including CUT3R~\cite{cut3r} and
TTT3R~\cite{ttt3r}. Results on VGGT-$\Omega$ and
Spark3R+VGGT-$\Omega$ are reported in dedicated rows of each table. All experiments run on a single NVIDIA H20
GPU in bfloat16, except for prediction heads that involve
SVD, which use float32 for numerical stability.

\subsection{Point Map Estimation}
\label{sec::experiment:3drecon}

\begin{table*}[htbp]
\centering
\caption{Evaluation of Point Map Estimation on 7-Scenes~\cite{7scenes} and NRGBD~\cite{nrgbd}.
\colorbox{red!30}{Red} and \colorbox{orange!30}{Orange} backgrounds indicate the best and the second best results, respectively. VGGT-$\Omega$~\cite{vggt-omega} was released after our submission and is thus compared separately.}
\label{tab:rec}
\begin{tabular}{l c cc cc cc c cc cc cc}
\toprule
\multirow{3.5}{*}{Method} & \multicolumn{7}{c}{7-Scenes} & \multicolumn{7}{c}{NRGBD} \\
\cmidrule(lr){2-8} \cmidrule(lr){9-15}
 & \multirow{2}{*}{Time~(s)$\downarrow$} & \multicolumn{2}{c}{Acc$\downarrow$} & \multicolumn{2}{c}{Comp$\downarrow$} & \multicolumn{2}{c}{NC$\uparrow$} & \multirow{2}{*}{Time~(s)$\downarrow$} & \multicolumn{2}{c}{Acc$\downarrow$} & \multicolumn{2}{c}{Comp$\downarrow$} & \multicolumn{2}{c}{NC$\uparrow$} \\
\cmidrule(lr){3-4} \cmidrule(lr){5-6} \cmidrule(lr){7-8} \cmidrule(lr){10-11} \cmidrule(lr){12-13} \cmidrule(lr){14-15}
 & & Mean & Med. & Mean & Med. & Mean & Med. & & Mean & Med. & Mean & Med. & Mean & Med. \\ 
\midrule
CUT3R~\cite{cut3r}    & 69.5   & 0.197 & 0.126 & 0.074 & 0.021 & 0.536 & 0.552 & 70.1   & 0.415 & 0.310 & 0.197 & 0.080 & 0.551 & 0.576 \\
TTT3R~\cite{ttt3r}    & 69.2   & 0.103 & 0.061 & 0.056 & 0.024 & 0.566 & 0.599 & 70.3   & 0.262 & 0.181 & 0.117 & 0.040 & 0.571 & 0.604 \\
VGGT~\cite{vggt}      & 1102.7 & 0.047 & 0.018 & 0.035 & 0.013 & 0.601 & 0.656 & 1110.4 & 0.042 & 0.030 & 0.025 & 0.008 & 0.798 & 0.907 \\
DA3~\cite{depthanything3}
                      & 976.4  & \cellcolor{orange!30}0.015 & \cellcolor{red!30}0.006 & \cellcolor{orange!30}0.018 & \cellcolor{red!30}0.009 & 0.620 & 0.687 & 983.3  & \cellcolor{red!30}0.012 & \cellcolor{red!30}0.006 & \cellcolor{red!30}0.015 & \cellcolor{red!30}0.004 & \cellcolor{orange!30}0.898 & \cellcolor{red!30}0.979 \\
$\pi^3$~\cite{pi3}    & 830.2  & \cellcolor{red!30}0.012 & \cellcolor{red!30}0.006 & \cellcolor{red!30}0.016 & \cellcolor{orange!30}0.010 & \cellcolor{red!30}0.646 & \cellcolor{red!30}0.729 & 835.7  & \cellcolor{orange!30}0.014 & \cellcolor{orange!30}0.007 & \cellcolor{orange!30}0.016 & \cellcolor{orange!30}0.005 & 0.884 & \cellcolor{orange!30}0.974 \\
FastVGGT~\cite{fastvggt}
                      & 247.2  & 0.043 & 0.010 & 0.025 & \cellcolor{red!30}0.009 & 0.595 & 0.646 & 245.7  & 0.024 & 0.017 & 0.018 & 0.007 & 0.768 & 0.891 \\
ZipMap~\cite{zipmap}  & 46.2   & 0.019 & \cellcolor{orange!30}0.007 & 0.025 & \cellcolor{red!30}0.009 & 0.605 & 0.661 & 46.6   & 0.022 & 0.012 & \cellcolor{orange!30}0.016 & 0.006 & 0.766 & 0.891 \\
\addlinespace
Ours+VGGT             & \cellcolor{orange!30}40.7   & 0.017 & \cellcolor{orange!30}0.007 & 0.020 & \cellcolor{orange!30}0.010 & 0.617 & 0.681 & \cellcolor{orange!30}40.9   & \cellcolor{orange!30}0.014 & 0.009 & \cellcolor{orange!30}0.016 & \cellcolor{orange!30}0.005 & 0.863 & 0.964 \\
Ours+DA3              & 58.0   & \cellcolor{orange!30}0.015 & \cellcolor{red!30}0.006 & \cellcolor{orange!30}0.018 & \cellcolor{red!30}0.009 & 0.620 & 0.687 & 56.9   & \cellcolor{red!30}0.012 & \cellcolor{red!30}0.006 & \cellcolor{red!30}0.015 & \cellcolor{red!30}0.004 & \cellcolor{red!30}0.899 & \cellcolor{red!30}0.979 \\
Ours+$\pi^3$          & \cellcolor{red!30}34.1   & \cellcolor{red!30}0.012 & \cellcolor{red!30}0.006 & \cellcolor{red!30}0.016 & \cellcolor{orange!30}0.010 & \cellcolor{orange!30}0.644 & \cellcolor{orange!30}0.726 & \cellcolor{red!30}34.3   & \cellcolor{orange!30}0.014 & 0.008 & \cellcolor{orange!30}0.016 & \cellcolor{orange!30}0.005 & 0.880 & 0.971 \\
\midrule
\multicolumn{15}{l}{\textit{Extension to VGGT-$\Omega$ (released after our initial submission):}} \\
\midrule
VGGT-$\Omega$~\cite{vggt-omega}
                      & 505.3 & 0.015 & \cellcolor{red!30}0.007 & \cellcolor{red!30}0.014 & \cellcolor{red!30}0.007 & \cellcolor{red!30}0.608 & \cellcolor{red!30}0.667 & 508.4 & 0.015 & 0.009 & 0.008 & \cellcolor{red!30}0.005 & 0.829 & \cellcolor{red!30}0.948 \\
Ours+VGGT-$\Omega$    & \cellcolor{red!30}27.7 & \cellcolor{red!30}0.014 & \cellcolor{red!30}0.007 & \cellcolor{red!30}0.014 & \cellcolor{red!30}0.007 & 0.606 & 0.664 & \cellcolor{red!30}27.8 & \cellcolor{red!30}0.014 & \cellcolor{red!30}0.008 & \cellcolor{red!30}0.007 & \cellcolor{red!30}0.005 & \cellcolor{red!30}0.835 & \cellcolor{red!30}0.948 \\
\bottomrule
\end{tabular}
\end{table*}

We evaluate scene-level point map estimation on
7-Scenes~\cite{7scenes} and NRGBD~\cite{nrgbd} using three
standard metrics: Accuracy (Acc), Completeness (Comp), and
Normal Consistency (NC). Results are reported in
Tab.~\ref{tab:rec}.

\textbf{Comparison with unaccelerated baselines.}
Across the three base models, Spark3R reduces inference time
by $17{\times}$--$27{\times}$ on 7-Scenes and NRGBD. When applied to DA3 and $\pi^3$, Spark3R preserves all three
metrics to within negligible margins of their respective
baselines, confirming that asymmetric token reduction introduces
minimal degradation.
Notably, Spark3R+VGGT substantially \emph{improves} over
unaccelerated VGGT: on 7-Scenes, mean Accuracy improves from
0.047 to 0.017 and mean Completeness from 0.035 to 0.020.
We attribute this to the context reduction introduced by
lightweight key-value pruning: as the input sequence grows, VGGT's global attention must distribute a fixed attention budget
across all tokens, inevitably diluting per-token attention
mass. By compressing the key-value tokens, Spark3R keeps the
effective context in a regime where attention operates more
robustly, yielding sharper geometry.

\textbf{Comparison with other acceleration methods.}
Spark3R+VGGT runs faster than ZipMap on both 7-Scenes and
NRGBD while delivering higher accuracy, completeness, and
normal consistency. Spark3R+DA3 is slightly slower than
ZipMap due to DA3's heavy Dual-DPT head, yet outperforms
ZipMap on all three metrics across both datasets.
Spark3R+$\pi^3$ runs faster than ZipMap and surpasses it
on every metric.
Among other acceleration methods, FastVGGT is roughly
$6{\times}$ slower than Spark3R+VGGT and yields lower
reconstruction quality, while the streaming methods CUT3R
and TTT3R are slower than every Spark3R variant and
exhibit substantially lower reconstruction quality owing
to the absence of global cross-frame reasoning.

\textbf{Extension to VGGT-$\bm{\Omega}$.}
On VGGT-$\Omega$, Spark3R
delivers an $18{\times}$ speedup while matching the reconstruction
quality of unaccelerated VGGT-$\Omega$ on both datasets,
confirming that Spark3R generalizes to newer feed-forward
architectures.

\subsection{Camera Pose Estimation}
\label{sec::experiment:pose_est}

\begin{table*}[!t]
\centering
\caption{Evaluation of Camera Pose Estimation on TUM-dynamics~\cite{tumdynamic}, ScanNet~\cite{scannet}, and Sintel~\cite{sintel}. \colorbox{red!30}{Red} and \colorbox{orange!30}{Orange} backgrounds indicate the best and the second best results, respectively. VGGT-$\Omega$~\cite{vggt-omega} was released after our submission and is thus compared separately.}
\label{tab:pose}
\resizebox{\textwidth}{!}{
\begin{tabular}{l cccc cccc cccc}
\toprule
Method & \multicolumn{4}{c}{TUM-dynamics} & \multicolumn{4}{c}{ScanNet} & \multicolumn{4}{c}{Sintel} \\
\cmidrule(lr){2-5} \cmidrule(lr){6-9} \cmidrule(lr){10-13}
& Time (s)$\downarrow$ & ATE$\downarrow$ & RPE$_t$$\downarrow$ & RPE$_r$$\downarrow$ & Time (s)$\downarrow$ & ATE$\downarrow$ & RPE$_t$$\downarrow$ & RPE$_r$$\downarrow$ & Time (s)$\downarrow$ & ATE$\downarrow$ & RPE$_t$$\downarrow$ & RPE$_r$$\downarrow$ \\
\midrule
CUT3R~\cite{cut3r}        & 81.6   & 0.165 & \cellcolor{red!30}0.007 & 0.534 & 86.5   & 0.783 & 0.021 & 0.841 & 4.1 & 0.209 & 0.074 & 0.638 \\
TTT3R~\cite{ttt3r}        & 81.4   & 0.110 & 0.009 & 0.451 & 87.1   & 0.424 & 0.021 & 0.595 & 4.1 & 0.208 & 0.093 & 0.734 \\
VGGT~\cite{vggt}          & 993.7  & \cellcolor{orange!30}0.018 & 0.009 & \cellcolor{orange!30}0.294 & 1162.7 & 0.156 & 0.055 & 1.532 & 2.0 & 0.172 & 0.062 & 0.469 \\
DA3~\cite{depthanything3} & 870.8  & \cellcolor{orange!30}0.018 & 0.009 & 0.295 & 1018.1 & \cellcolor{orange!30}0.057 & 0.013 & 0.350 & 2.6 & 0.102 & 0.053 & 0.441 \\
$\pi^3$~\cite{pi3}        & 739.4  & 0.022 & \cellcolor{orange!30}0.008 & 0.295 & 865.1  & \cellcolor{red!30}0.056 & \cellcolor{red!30}0.011 & \cellcolor{red!30}0.264 & \cellcolor{orange!30}1.6 & \cellcolor{red!30}0.073 & \cellcolor{red!30}0.038 & \cellcolor{red!30}0.294 \\
FastVGGT~\cite{fastvggt}  & 253.8  & 0.020 & 0.010 & 0.297 & 256.8  & 0.078 & 0.025 & 0.565 & 2.4 & 0.167 & 0.070 & 0.522 \\
ZipMap~\cite{zipmap}      & 44.6   & 0.021 & 0.010 & 0.314 & 43.6   & 0.066 & 0.017 & 0.361 & 3.9 & 0.134 & 0.067 & 0.429 \\
\addlinespace
Ours+VGGT                 & \cellcolor{orange!30}38.4 & \cellcolor{red!30}0.016 & 0.010 & 0.297 & \cellcolor{orange!30}41.3 & 0.065 & 0.019 & 0.403 & \cellcolor{orange!30}1.6 & 0.172 & 0.058 & 0.478 \\
Ours+DA3                  & 54.0   & \cellcolor{red!30}0.016 & \cellcolor{orange!30}0.008 & \cellcolor{red!30}0.291 & 57.8 & \cellcolor{orange!30}0.057 & \cellcolor{orange!30}0.012 & 0.320 & 2.6 & \cellcolor{orange!30}0.100 & \cellcolor{orange!30}0.050 & 0.431 \\
Ours+$\pi^3$              & \cellcolor{red!30}32.2 & 0.022 & \cellcolor{orange!30}0.008 & 0.298 & \cellcolor{red!30}34.6 & 0.058 & \cellcolor{orange!30}0.012 & \cellcolor{orange!30}0.280 & \cellcolor{red!30}1.4 & \cellcolor{red!30}0.073 & \cellcolor{red!30}0.038 & \cellcolor{orange!30}0.295 \\
\midrule
\multicolumn{13}{l}{\textit{Extension to VGGT-$\Omega$ (released after our initial submission):}} \\
\midrule
VGGT-$\Omega$~\cite{vggt-omega}
                          & 448.7 & \cellcolor{red!30}0.008 & \cellcolor{red!30}0.005 & \cellcolor{red!30}0.263 & 525.1 & 0.067 & \cellcolor{red!30}0.018 & 0.463 & 1.3 & \cellcolor{red!30}0.039 & \cellcolor{red!30}0.026 & 0.238 \\
Ours+VGGT-$\Omega$        & \cellcolor{red!30}28.7 & 0.009 & 0.006 & 0.278 & \cellcolor{red!30}28.0 & \cellcolor{red!30}0.063 & 0.019 & \cellcolor{red!30}0.426 & \cellcolor{red!30}1.2 & \cellcolor{red!30}0.039 & 0.027 & \cellcolor{red!30}0.236 \\
\bottomrule
\end{tabular}
}
\end{table*}

We evaluate camera pose estimation on
TUM-dynamics~\cite{tumdynamic}, ScanNet~\cite{scannet}, and
Sintel~\cite{sintel}, reporting the Absolute Trajectory Error
(ATE) and the Relative Pose Error for translation (RPE$_t$)
and rotation (RPE$_r$), all computed after
$\mathrm{Sim}(3)$ alignment. Results are presented in
Tab.~\ref{tab:pose}.

\textbf{Comparison with unaccelerated baselines.}
On the two long-sequence datasets (TUM-dynamics and ScanNet),
Spark3R delivers up to $28{\times}$ speedup over the
corresponding baselines without sacrificing pose accuracy.
When applied to DA3 and $\pi^3$, it matches the respective
baselines to within negligible margins on all metrics.
Spark3R+VGGT again substantially \emph{improves} over
unaccelerated VGGT on ScanNet, with ATE dropping from 0.156
to 0.065, consistent with the attention-dilution effect
discussed in Sec.~\ref{sec::experiment:3drecon}.

On Sintel, sequences contain at most 50 frames, well below
the regime where global attention becomes a bottleneck.
Spark3R's length-adaptive schedule accordingly sets both
base reduction factors to $r_\text{Q}{=}1$ and
$r_\text{KV}{=}1$, yet the layer-adaptive schedule remains
active: low-sensitivity layers still receive an elevated
key-value reduction factor of $3\,r_\text{KV}{=}3$. The
matched pose accuracy on these short sequences confirms the
standalone effectiveness of the layer-adaptive mechanism.

\textbf{Comparison with other acceleration methods.}
Spark3R+VGGT runs faster than ZipMap on both TUM-dynamics
and ScanNet while attaining comparable pose accuracy.
Spark3R+DA3 is again slightly slower than ZipMap but
outperforms it on all three pose metrics, and Spark3R+$\pi^3$
runs faster than ZipMap while surpassing it on every metric.
Among other acceleration methods, FastVGGT remains over
$6{\times}$ slower than Spark3R+VGGT at comparable or
lower pose accuracy, while the streaming methods CUT3R and
TTT3R are slower than every Spark3R variant and exhibit
substantially lower pose accuracy on ATE and RPE$_r$.

\textbf{Extension to VGGT-$\bm{\Omega}$.}
On VGGT-$\Omega$, Spark3R delivers a $16{\times}$ speedup on
TUM-dynamics and a $19{\times}$ speedup on ScanNet while
matching the pose accuracy of unaccelerated VGGT-$\Omega$
on ScanNet and Sintel and trailing it slightly on
TUM-dynamics, further confirming the
generality of Spark3R across feed-forward backbones.

\subsection{Video Depth Estimation}
\label{sec::experiment:depth_est}

\begin{table*}[htbp]
\centering
\caption{Evaluation of Video Depth Estimation on Bonn~\cite{bonn}, KITTI~\cite{kitti}, and Sintel~\cite{sintel}. \colorbox{red!30}{Red} and \colorbox{orange!30}{Orange} backgrounds indicate the best and the second best results, respectively. VGGT-$\Omega$~\cite{vggt-omega} was released after our submission and is thus compared separately.}
\label{tab:depth_complete}
\resizebox{\textwidth}{!}{
\begin{tabular}{l ccc ccc ccc}
\toprule
Method & \multicolumn{3}{c}{Bonn} & \multicolumn{3}{c}{KITTI} & \multicolumn{3}{c}{Sintel} \\
\cmidrule(lr){2-4} \cmidrule(lr){5-7} \cmidrule(lr){8-10}
& Time (s)$\downarrow$ & Abs Rel$\downarrow$ & $\delta < 1.25$$\uparrow$ & Time (s)$\downarrow$ & Abs Rel$\downarrow$ & $\delta < 1.25$$\uparrow$ & Time (s)$\downarrow$ & Abs Rel$\downarrow$ & $\delta < 1.25$$\uparrow$ \\
\midrule
CUT3R~\cite{cut3r}        & 58.3  & 0.072 & 95.4 & 21.6 & 0.117 & 87.8 & 4.2 & 0.680 & 52.7 \\
TTT3R~\cite{ttt3r}        & 58.5  & 0.060 & 97.4 & 21.5 & 0.108 & 89.3 & 4.1 & 0.635 & 55.1 \\
VGGT~\cite{vggt}          & 552.4 & 0.046 & 98.2 & 31.5 & 0.078 & 94.2 & 2.1 & 0.263 & 64.6 \\
DA3~\cite{depthanything3} & 493.4 & 0.040 & 98.2 & 28.8 & 0.063 & 96.6 & 2.6 & 0.238 & 65.6 \\
$\pi^3$~\cite{pi3}        & 413.1 & \cellcolor{red!30}0.033 & \cellcolor{red!30}98.9 & 22.6 & \cellcolor{red!30}0.046 & \cellcolor{red!30}97.9 & \cellcolor{orange!30}1.6 & \cellcolor{orange!30}0.217 & \cellcolor{orange!30}71.8 \\
FastVGGT~\cite{fastvggt}  & 182.8 & 0.043 & 98.4 & 11.2 & 0.078 & 94.2 & 2.3 & 0.287 & 64.2 \\
ZipMap~\cite{zipmap}      & 32.8  & 0.040 & \cellcolor{orange!30}98.5 & 7.5  & 0.065 & 95.2 & 3.8 & 0.222 & 68.9 \\
\addlinespace
Ours+VGGT                 & \cellcolor{orange!30}31.1 & 0.044 & 98.1 & \cellcolor{orange!30}5.3 & 0.078 & 94.0 & \cellcolor{orange!30}1.8 & 0.270 & 64.6 \\
Ours+DA3                  & 38.3 & 0.039 & 98.3 & 7.2 & 0.064 & 96.4 & 2.5 & 0.240 & 65.5 \\
Ours+$\pi^3$              & \cellcolor{red!30}23.3 & \cellcolor{orange!30}0.035 & \cellcolor{red!30}98.9 & \cellcolor{red!30}3.9 & \cellcolor{orange!30}0.047 & \cellcolor{orange!30}97.8 & \cellcolor{red!30}1.4 & \cellcolor{red!30}0.216 & \cellcolor{red!30}71.9 \\
\midrule
\multicolumn{10}{l}{\textit{Extension to VGGT-$\Omega$ (released after our intial submission):}} \\
\midrule
VGGT-$\Omega$~\cite{vggt-omega}
                          & 252.8 & 0.036 & 98.3 & 16.6 & \cellcolor{red!30}0.064 & \cellcolor{red!30}96.6 & 1.3 & \cellcolor{red!30}0.091 & \cellcolor{red!30}92.6 \\
Ours+VGGT-$\Omega$        & \cellcolor{red!30}19.2 & \cellcolor{red!30}0.035 & \cellcolor{red!30}98.4 & \cellcolor{red!30}3.6 & 0.067 & 96.0 & \cellcolor{red!30}1.2 & \cellcolor{red!30}0.091 & \cellcolor{red!30}92.6 \\
\bottomrule
\end{tabular}
}
\end{table*}

We evaluate video depth estimation on Bonn~\cite{bonn},
KITTI~\cite{kitti}, and Sintel~\cite{sintel}, covering
indoor, outdoor, and synthetic scenes respectively. We
adopt Absolute Relative Error (Abs Rel) and the threshold
accuracy $\delta < 1.25$ as evaluation metrics, with
predicted depth maps aligned to the ground truth using a
per-sequence scale and shift. Results are reported in
Tab.~\ref{tab:depth_complete}.

\textbf{Comparison with unaccelerated baselines.}
Across the three base models, Spark3R achieves
$13{\times}$--$18{\times}$ speedup on Bonn (avg.\ 623
frames) and $4{\times}$--$6{\times}$ on KITTI (avg.\ 264
frames) at the cost of minor depth degradation: Abs Rel
increases by at most 0.002 and $\delta < 1.25$ drops by
at most 0.2. On Sintel (avg.\ 46 frames), Spark3R closely
matches the depth accuracy of every base model, further
confirming the standalone benefit of the layer-adaptive
mechanism noted in Sec.~\ref{sec::experiment:pose_est}.

\textbf{Comparison with other acceleration methods.}
Spark3R+VGGT runs faster than ZipMap on all three
datasets but trails it on depth metrics. Spark3R+DA3 is
slightly slower than ZipMap on Bonn, while achieving
comparable depth accuracy on Bonn and slightly better
accuracy on KITTI. On Sintel, ZipMap outperforms both
Spark3R+VGGT and Spark3R+DA3 in Abs Rel and
$\delta < 1.25$. Spark3R+$\pi^3$
runs faster than ZipMap on every dataset and surpasses it
on every depth metric. Among other acceleration methods,
FastVGGT is roughly $2{\times}$--$6{\times}$ slower than
Spark3R+VGGT across Bonn and KITTI while providing
slightly better depth accuracy. The streaming methods
CUT3R and TTT3R are slower than every Spark3R variant
and exhibit substantially lower depth accuracy on all
three datasets.

\textbf{Extension to VGGT-$\bm{\Omega}$.}
On VGGT-$\Omega$, Spark3R delivers a $13{\times}$ speedup
on Bonn and a $4.6{\times}$ speedup on KITTI, matching the
depth accuracy of unaccelerated VGGT-$\Omega$ on Bonn
and trailing it slightly on KITTI. Sintel highlights the
benefit most clearly: VGGT-$\Omega$ alone obtains an Abs
Rel of $0.091$, far ahead of every other base model
($0.217$--$0.287$), and Spark3R retains this advantage
while becoming the fastest variant on every benchmark in
Tab.~\ref{tab:depth_complete}.
These results further confirm that Spark3R generalizes
across feed-forward backbones.

\begin{figure*}[ht]
\vspace{0pt}
\includegraphics[width=\textwidth]{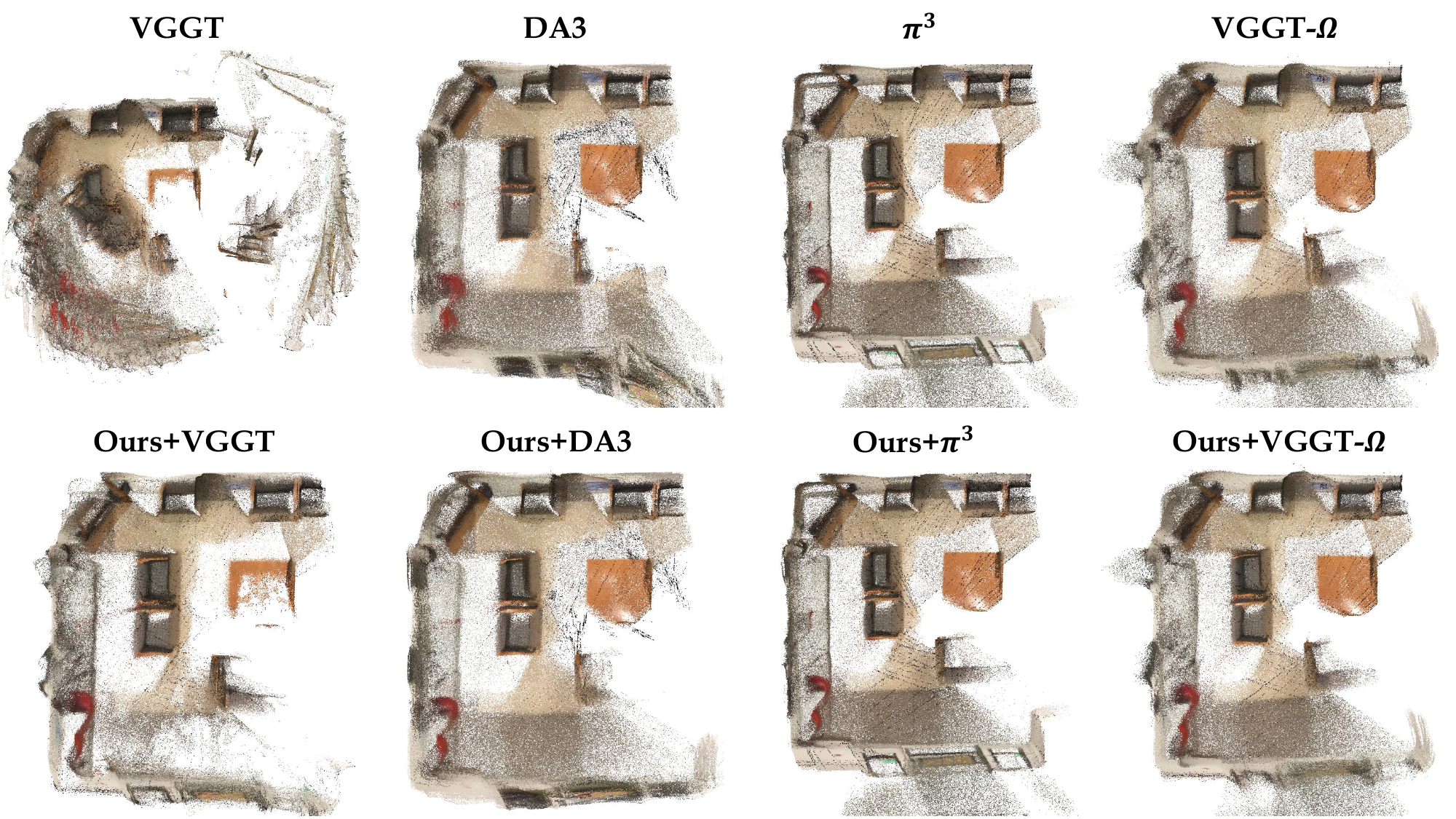}
\vspace{-10pt}
\caption{\textbf{Qualitative comparison with unaccelerated
  baselines.} Each pair shows the original model and its Spark3R-accelerated counterpart. 
  Spark3R preserves fine-grained geometric details and produces point clouds visually comparable to the
  unaccelerated baselines. Notably,
for VGGT, Spark3R even improves the reconstruction quality
by alleviating attention dilution on long sequences.}
\label{fig:qual_baseline}
\end{figure*}

\begin{figure*}[ht]
\vspace{0pt}
\includegraphics[width=\textwidth]{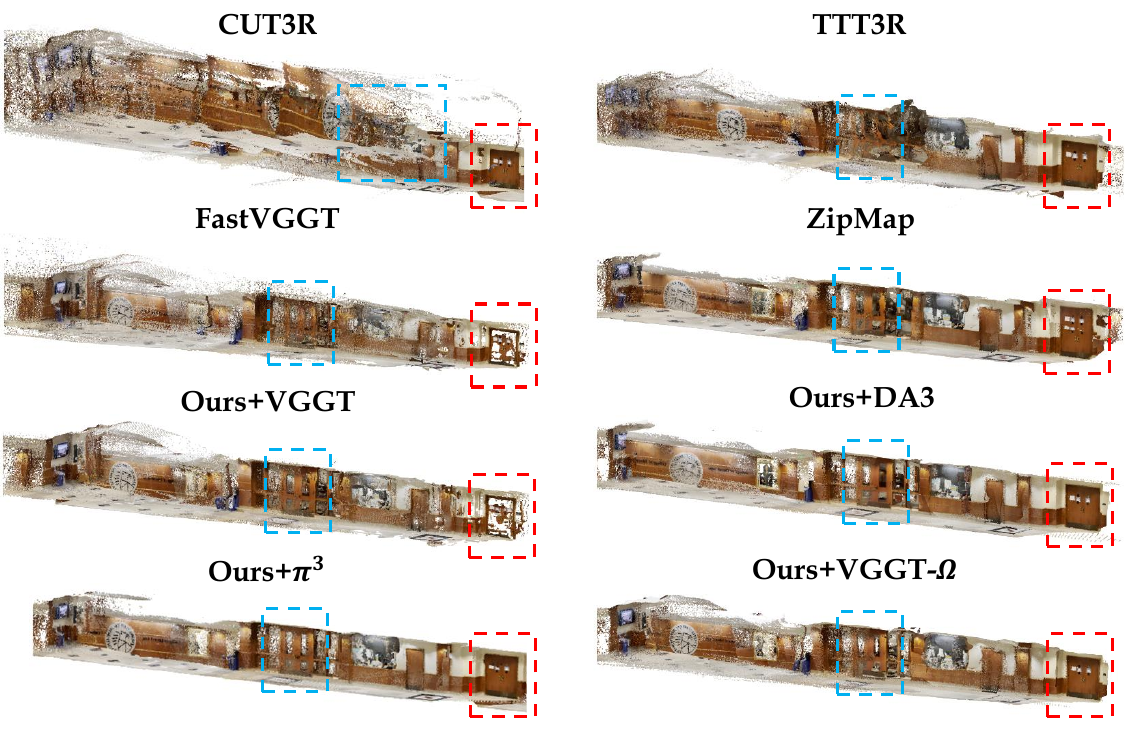}
\vspace{-10pt}
\caption{\textbf{Qualitative comparison with other
acceleration methods.} The streaming methods CUT3R and
TTT3R produce fragmented and noisy reconstructions.
FastVGGT produces blurred geometry in the blue dashed box,
while ZipMap exhibits subtle structural distortions in the
red dashed box (\ie, misaligned door parts).
Spark3R+VGGT sharpens the reconstruction over FastVGGT,
and Spark3R applied to $\pi^3$, DA3, and VGGT-$\Omega$
further surpasses ZipMap with geometrically faithful outputs.}
\label{fig:qual_others}
\end{figure*}

\subsection{Qualitative Results}
\label{sec:qual}
Figures~\ref{fig:qual_baseline} and~\ref{fig:qual_others}
present qualitative point-cloud comparisons. In
Figure~\ref{fig:qual_baseline}, Spark3R applied to $\pi^3$ and VGGT-$\Omega$
produces point clouds virtually identical to the
unaccelerated baselines. For DA3, the original reconstruction
exhibits a visible geometric misalignment in the bottom-right
region, which we attribute to accumulated pose drift; Spark3R+DA3
corrects this misalignment, consistent with the improved
rotational pose accuracy reported in
Tab.~\ref{tab:pose}. The effect is most pronounced for
VGGT: the baseline suffers from severe pose drift and large-scale geometric distortions. By mitigating the attention-dilution
effect discussed in Sec.~\ref{sec::experiment:3drecon},
Spark3R restores coherent pose estimates and substantially
improves the reconstruction.

In Figure~\ref{fig:qual_others}, the streaming methods
CUT3R and TTT3R produce visibly degraded reconstructions,
with CUT3R's geometry largely fragmented and TTT3R
retaining coarser structure but still showing prominent
artifacts. FastVGGT produces noticeably blurred geometry
in the blue dashed box; Spark3R+VGGT alleviates this
blurriness, recovering sharper textures and cleaner
structural details. ZipMap delivers a more complete
reconstruction overall, yet still exhibits subtle
geometric distortions---\ie, the upper and middle portions
of the door in the red dashed box are visibly misaligned.
Spark3R applied to $\pi^3$, DA3, and VGGT-$\Omega$ eliminates these
distortions, producing sharp and structurally faithful
reconstructions.

\subsection{Ablation Studies}
\label{sec:ablation}

We conduct ablation studies on VGGT to validate each design
choice in Spark3R. Unless otherwise noted, all experiments
use 256-frame sequences from the 7-Scenes~\cite{7scenes} dataset, except
for the group size ablation, which uses 512-frame sequences
to more clearly expose the quadratic cost of global matching
on longer inputs.

\begin{figure}[htbp]
  \centering
  \includegraphics[width=\linewidth]{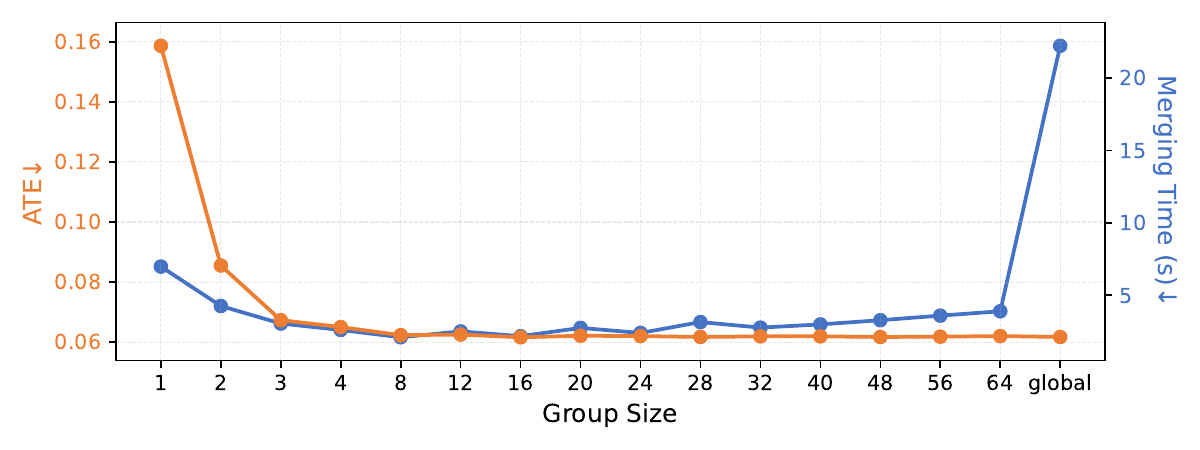}
  \vspace{-16pt}
  \caption{ATE and wall-clock merging time as a function of
  the group size~$G$ on 512-frame sequences from 7-Scenes~\cite{7scenes}.}
  \label{fig:qm_groupsize}
\end{figure}
\textbf{Group Size for Intra-Group Query Merging.}
Figure~\ref{fig:qm_groupsize} plots ATE and wall-clock
merging time as a function of the group size~$G$. ATE drops
sharply from $G{=}1$ to $G{=}8$ and plateaus for
$G \geq 8$, confirming that beneficial merging pairs are
concentrated within a small temporal neighborhood,
consistent with the locality analysis in
Figure~\ref{fig:merge_distance}. Merging time remains low
and roughly constant for moderate group sizes, as the
intra-group matching cost scales as
$\mathcal{O}(S \cdot G)$, linear in the sequence
length~$S$ for fixed~$G$. At the global setting where
$G{=}S$, however, the cost reverts to $\mathcal{O}(S^2)$:
merging time exceeds 20\,s on 512-frame inputs, becoming a
significant fraction of total inference time. Any group size
in the range of roughly 8 to 64 achieves near-optimal ATE
with low matching overhead; we adopt $G{=}20$ as our default
throughout all other experiments.


\begin{table}[htbp]
\centering
\caption{Ablation on the query reduction factor
$r_\text{Q}$. \textbf{Bold} indicates the best result and \underline{underline} indicates the second best.}
\label{tab:ablation_qr}
\begin{tabular}{l cccc}
\toprule
$r_{\text{Q}}$ & Time~(s)$\downarrow$ & ATE$\downarrow$ & RPE$_{t}$$\downarrow$ & RPE$_{r}$$\downarrow$ \\
\midrule
1  & 89.9           & \textbf{0.0291} & \textbf{0.0081} & \textbf{0.2656} \\
2  & 50.4           & \textbf{0.0291} & \underline{0.0085} & \underline{0.2713} \\
3  & 37.0           & \underline{0.0292} & 0.0087          & 0.2765          \\
4  & 30.5           & 0.0293          & 0.0091          & 0.2811          \\
8  & 20.8           & 0.0298          & 0.0097          & 0.2977          \\
12 & \underline{17.1} & 0.0305          & 0.0105          & 0.3190          \\
16 & \textbf{15.4}  & 0.0310          & 0.0111          & 0.3397          \\
\bottomrule
\end{tabular}
\end{table}
\textbf{Query Reduction Factor.}
Tab.~\ref{tab:ablation_qr} provides a detailed numerical
breakdown of the query-token sensitivity introduced in
Figure~\ref{fig:qkv_sensitivity}, sweeping $r_\text{Q}$
with $r_\text{KV}$ fixed at~1 to isolate the effect of
query compression. ATE remains nearly constant from
$r_\text{Q}{=}1$ through $r_\text{Q}{=}4$, varying only
between 0.0291 and 0.0293, while RPE metrics exhibit a mild
but steady increase, with RPE$_r$ rising from 0.2656 to
0.2811. Meanwhile, inference time drops from 89.9\,s to
30.5\,s. Beyond $r_\text{Q}{=}4$,
degradation becomes more apparent: ATE rises to 0.0298 at
$r_\text{Q}{=}8$ and to 0.0310 at $r_\text{Q}{=}16$, with
RPE metrics following the same trend. These results
delineate a clear safe operating range for query
compression, and the length-adaptive schedule in
Eq.~\eqref{eq:adaptive_rq}, which caps $r_\text{Q}$ at~4,
is designed to stay well within this regime.

\begin{figure}[htbp]
  \centering
  \vspace{-16pt}
  \includegraphics[width=\linewidth]{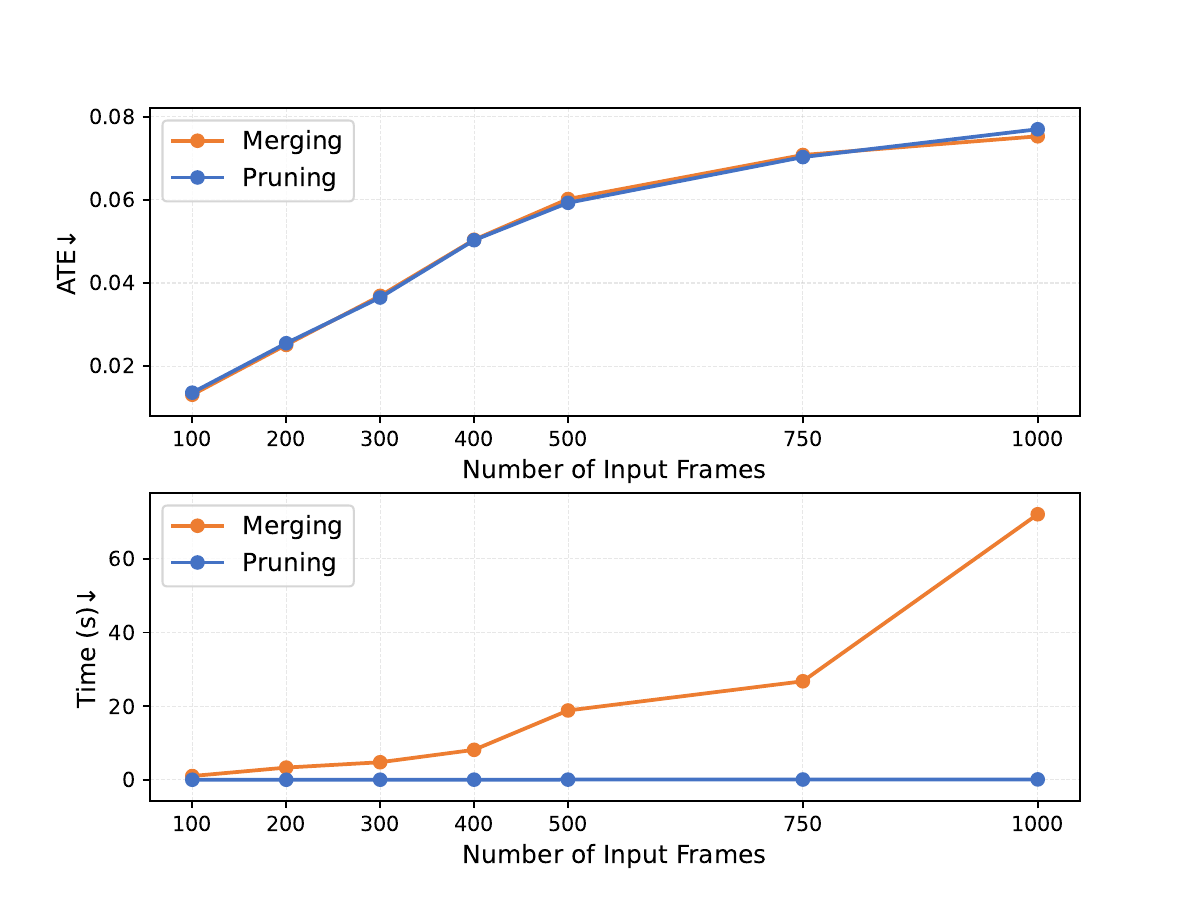}
  \vspace{-16pt}
  \caption{\textbf{Merging vs.\ pruning for key-value
  tokens.} Both strategies use the same temporal stride
  partitioning into source and destination tokens.
  \textit{Top}: ATE as a function of the number of input
  frames; both achieve nearly identical pose error.
  \textit{Bottom}: wall-clock token reduction time. Token merging
  grows superlinearly due to the bipartite similarity
  computation, while token pruning remains near zero
  throughout.}
  \label{fig:kv_update}
\end{figure}
\textbf{Merging vs.\ Pruning for Key-Value Tokens.}
Figure~\ref{fig:kv_update} compares two key-value
compression strategies that share the same source/destination
partitioning via temporal stride sampling but differ in how
source tokens are handled. Token merging computes bipartite
cosine similarities between all source tokens and destination tokens
to assign each source token to its most similar destination, then averages the matched pairs to
update the destination representation. Token pruning simply
discards all source tokens and retains the destination
tokens unchanged.
The top panel shows that both strategies yield nearly
identical ATE across all tested input lengths from 100 to
1{,}000 frames, confirming that the averaging step provides
minimal benefit for key-value tokens, as expected from
the high source-to-destination similarity observed in
Figure~\ref{fig:kv_sim}.
The bottom panel reveals the critical practical difference:
the wall-clock time for token merging grows
superlinearly with the number of input frames due to the
bipartite similarity computation between source and
destination tokens, exceeding 70\,s at
1{,}000 frames, whereas token pruning incurs effectively
zero overhead regardless of sequence length. Since the identity closely approximates the averaging update, replacing
merging with pruning eliminates the dominant matching cost
without sacrificing reconstruction quality.


\begin{table}[htbp]
\centering
\caption{Ablation on the key-value reduction factor
$r_\text{KV}$. \textbf{Bold} indicates the best result and \underline{underline} indicates the second best.}
\label{tab:ablation_kvr}
\begin{tabular}{l cccc}
\toprule
$r_\text{KV}$ & Time~(s)$\downarrow$ & ATE$\downarrow$ & RPE$_{t}$$\downarrow$ & RPE$_{r}$$\downarrow$ \\
\midrule
1  & 89.9           & 0.0291          & \textbf{0.0081} & 0.2656 \\
2  & 49.8           & \textbf{0.0289} & \textbf{0.0081} & 0.2652 \\
3  & 36.3           & \underline{0.0290} & \textbf{0.0081} & \textbf{0.2635} \\
4  & 29.6           & \underline{0.0290} & \underline{0.0082} & \underline{0.2651} \\
8  & 19.4           & \textbf{0.0289} & 0.0085          & 0.2661 \\
16 & \underline{14.5} & 0.0293          & 0.0088          & 0.2696 \\
32 & \textbf{12.0}  & 0.0302          & 0.0097          & 0.2810 \\
\bottomrule
\end{tabular}
\end{table}
\textbf{Key-Value Reduction Factor.}
Tab.~\ref{tab:ablation_kvr} provides a detailed numerical
breakdown of the key-value token sensitivity introduced in
Figure~\ref{fig:qkv_sensitivity}, sweeping $r_\text{KV}$
with $r_\text{Q}$ fixed at~1. Pose error remains virtually
flat from $r_\text{KV}{=}1$ through $r_\text{KV}{=}8$, with
ATE fluctuating between 0.0289 and 0.0290, while inference
time drops from 89.9\,s to 19.4\,s. Even at
$r_\text{KV}{=}16$, ATE rises only marginally to 0.0293.
Noticeable degradation appears only at
$r_\text{KV}{=}32$, where ATE reaches 0.0302. This high
tolerance to key-value compression confirms the wide safe
operating range for key-value tokens and supports the
length-adaptive schedule in Eq.~\eqref{eq:adaptive_rkv}, which
yields $r_\text{KV}{=}7$ for 256-frame inputs, well within
the quality-preserving regime.


\begin{table}[htbp]
\centering
\caption{Ablation on the layer-adaptive multiplier~$l$.
\textbf{Bold} indicates the best result and \underline{underline} indicates the second best.}
\label{tab:ablation_layer}
\begin{tabular}{l cccc}
\toprule
$l$ & Time (s) $\downarrow$ & ATE $\downarrow$ & RPE$_t$ $\downarrow$ & RPE$_r$ $\downarrow$ \\
\midrule
1 & 19.6           & \textbf{0.0289} & \underline{0.0085} & \underline{0.2661} \\
2 & 15.8           & \textbf{0.0289} & \textbf{0.0084}    & \textbf{0.2660} \\
3 & 14.4           & \underline{0.0292} & \textbf{0.0084}    & 0.2668          \\
4 & 13.7           & 0.0293          & \textbf{0.0084}    & 0.2671          \\
6 & \underline{13.3} & 0.0294          & \underline{0.0085} & 0.2693          \\
8 & \textbf{12.8}  & 0.0302          & \underline{0.0085} & 0.2679          \\
\bottomrule
\end{tabular}
\end{table}
\textbf{Layer-Adaptive Key-Value Reduction Schedule.}
Tab.~\ref{tab:ablation_layer} fixes $r_\text{Q}{=}1$ and
$r_\text{KV}{=}8$ and evaluates the layer-adaptive schedule
by varying the multiplier~$l$ applied to the base
key-value reduction factor $r_\text{KV}$ for low-sensitivity layers, while
high-sensitivity layers retain the base
reduction factor. At $l{=}1$, all layers use the same
reduction factor and inference takes 19.6\,s. Increasing $l$
to~2 reduces time to 15.8\,s with no performance degradation, suggesting that low-sensitivity layers
tolerate more aggressive pruning without losing useful
information. Further increases
continue to reduce inference time, reaching 12.8\,s at
$l{=}8$, but ATE begins to rise beyond $l{=}3$. We adopt
$l{=}3$ as the default setting of Spark3R, as it
provides a favorable quality--efficiency trade-off.

\section{Conclusion}
\label{sec:conclusion}

We introduced Spark3R, a training-free, plug-and-play
acceleration framework for feed-forward 3D reconstruction
models. 
Spark3R applies conservative intra-group token merging for query tokens
and aggressive lightweight token pruning for key-value tokens, and further integrates a layer-adaptive key-value reduction schedule.
Spark3R integrates directly into VGGT, $\pi^3$,
Depth-Anything-3, and VGGT-$\Omega$, achieving up to $28{\times}$ speedup on
1{,}000-frame inputs while maintaining competitive
reconstruction quality across diverse benchmarks.

\textbf{Limitations and Future Work.}
As a training-free framework, Spark3R inherits the 
quality ceiling of its base model: 
it can preserve or improve predictions by alleviating attention dilution, but cannot
compensate for intrinsic failure modes. Jointly fine-tuning
the base model with Spark3R so that it adapts to the
compressed token sequences is a promising direction for
future work.

\bibliographystyle{IEEEtran}
\bibliography{reference}

@misc{attn,
      title={Attention Is All You Need}, 
      author={Ashish Vaswani and Noam Shazeer and Niki Parmar and Jakob Uszkoreit and Llion Jones and Aidan N. Gomez and Lukasz Kaiser and Illia Polosukhin},
      year={2023},
      eprint={1706.03762},
      archivePrefix={arXiv},
      primaryClass={cs.CL},
      url={https://arxiv.org/abs/1706.03762}, 
}

@inproceedings{flash-attn,
  title={Flash{A}ttention: Fast and Memory-Efficient Exact Attention with {IO}-Awareness},
  author={Dao, Tri and Fu, Daniel Y. and Ermon, Stefano and Rudra, Atri and R{\'e}, Christopher},
  booktitle={Proc. of NeurIPS},
  year={2022}
}

@misc{fastkv,
      title={Fast KV Compaction via Attention Matching}, 
      author={Adam Zweiger and Xinghong Fu and Han Guo and Yoon Kim},
      year={2026},
      eprint={2602.16284},
      archivePrefix={arXiv},
      primaryClass={cs.LG},
      url={https://arxiv.org/abs/2602.16284}, 
}

@inproceedings{vggt,
  title={VGGT: Visual Geometry Grounded Transformer},
  author={Wang, Jianyuan and Chen, Minghao and Karaev, Nikita and Vedaldi, Andrea and Rupprecht, Christian and Novotny, David},
  booktitle={Proc. of CVPR},
  year={2025}
}

@misc{vggt3,
      title={VGG-T$^3$: Offline Feed-Forward 3D Reconstruction at Scale}, 
      author={Sven Elflein and Ruilong Li and Sérgio Agostinho and Zan Gojcic and Laura Leal-Taixé and Qunjie Zhou and Aljosa Osep},
      year={2026},
      eprint={2602.23361},
      archivePrefix={arXiv},
      primaryClass={cs.CV},
      url={https://arxiv.org/abs/2602.23361}, 
}

@misc{zipmap,
      title={ZipMap: Linear-Time Stateful 3D Reconstruction via Test-Time Training}, 
      author={Haian Jin and Rundi Wu and Tianyuan Zhang and Ruiqi Gao and Jonathan T. Barron and Noah Snavely and Aleksander Holynski},
      year={2026},
      eprint={2603.04385},
      archivePrefix={arXiv},
      primaryClass={cs.CV},
      url={https://arxiv.org/abs/2603.04385}, 
}

@misc{lact,
      title={Test-Time Training Done Right}, 
      author={Tianyuan Zhang and Sai Bi and Yicong Hong and Kai Zhang and Fujun Luan and Songlin Yang and Kalyan Sunkavalli and William T. Freeman and Hao Tan},
      year={2025},
      eprint={2505.23884},
      archivePrefix={arXiv},
      primaryClass={cs.LG},
      url={https://arxiv.org/abs/2505.23884}, 
}

@article{pi3,
  title={$\pi^3$: Scalable Permutation-Equivariant Visual Geometry Learning}, 
  author={Wang, Yifan and Zhou, Jianjun and Zhu, Haoyi and Chang, Wenzheng and Zhou, Yang and Li, Zizun and Chen, Junyi and Pang, Jiangmiao and Shen, Chunhua and He, Tong},
  journal={arXiv preprint arXiv:2507.13347},
  year={2025}
}

@article{depthanything3,
  title={Depth Anything 3: Recovering the visual space from any views},
  author={Haotong Lin and Sili Chen and Jun Hao Liew and Donny Y. Chen and Zhenyu Li and Guang Shi and Jiashi Feng and Bingyi Kang},
  journal={arXiv preprint arXiv:2511.10647},
  year={2025}
}

@inproceedings{sfm,
  title={Structure-from-motion revisited},
  author={Schonberger, Johannes L and Frahm, Jan-Michael},
  booktitle={Proc. of CVPR},
  pages={4104--4113},
  year={2016}
}

@article{photo,
author = {Snavely, Noah and Seitz, Steven M. and Szeliski, Richard},
title = {Photo tourism: exploring photo collections in 3D},
year = {2006},
issue_date = {July 2006},
publisher = {Association for Computing Machinery},
address = {New York, NY, USA},
volume = {25},
number = {3},
issn = {0730-0301},
url = {https://doi.org/10.1145/1141911.1141964},
doi = {10.1145/1141911.1141964},
journal = {ACM Trans. Graph.},
month = jul,
pages = {835–846},
numpages = {12}
}

@article{3dgs,
  title={3D Gaussian splatting for real-time radiance field rendering.},
  author={Kerbl, Bernhard and Kopanas, Georgios and Leimk{\"u}hler, Thomas and Drettakis, George},
  journal={{ACM Transactions on Graphics}},
  volume={42},
  number={4},
  pages={139--1},
  year={2023}
}

@inproceedings{mvs,
  title={Pixelwise view selection for unstructured multi-view stereo},
  author={Sch{\"o}nberger, Johannes L and Zheng, Enliang and Frahm, Jan-Michael and Pollefeys, Marc},
  booktitle={Proc. of ECCV},
  pages={501--518},
  year={2016},
  organization={Springer}
}

@inproceedings{dust3r,
  title={Dust3r: Geometric 3d vision made easy},
  author={Wang, Shuzhe and Leroy, Vincent and Cabon, Yohann and Chidlovskii, Boris and Revaud, Jerome},
  booktitle={Proc. of CVPR},
  pages={20697--20709},
  year={2024}
}

@inproceedings{mast3r,
  title={Grounding image matching in 3d with mast3r},
  author={Leroy, Vincent and Cabon, Yohann and Revaud, J{\'e}r{\^o}me},
  booktitle={{Proc. of ECCV}},
  pages={71--91},
  year={2024},
  organization={Springer}
}

@inproceedings{cut3r,
  title={Continuous 3d perception model with persistent state},
  author={Wang, Qianqian and Zhang, Yifei and Holynski, Aleksander and Efros, Alexei A and Kanazawa, Angjoo},
  booktitle={{Proc. of CVPR}},
  pages={10510--10522},
  year={2025}
}

@misc{point3r,
      title={Point3R: Streaming 3D Reconstruction with Explicit Spatial Pointer Memory}, 
      author={Yuqi Wu and Wenzhao Zheng and Jie Zhou and Jiwen Lu},
      year={2025},
      eprint={2507.02863},
      archivePrefix={arXiv},
      primaryClass={cs.CV},
      url={https://arxiv.org/abs/2507.02863}, 
}

@inproceedings{mvsnet,
  title={Mvsnet: Depth inference for unstructured multi-view stereo},
  author={Yao, Yao and Luo, Zixin and Li, Shiwei and Fang, Tian and Quan, Long},
  booktitle={{Proc. of ECCV}},
  pages={767--783},
  year={2018}
}

@article{nerf,
  title={Nerf: Representing scenes as neural radiance fields for view synthesis},
  author={Mildenhall, Ben and Srinivasan, Pratul P and Tancik, Matthew and Barron, Jonathan T and Ramamoorthi, Ravi and Ng, Ren},
  journal={{Communications of the ACM}},
  volume={65},
  number={1},
  pages={99--106},
  year={2021},
  publisher={ACM New York, NY, USA}
}

@article{fastvggt,
  title={Fastvggt: Training-free acceleration of visual geometry transformer},
  author={Shen, You and Zhang, Zhipeng and Qu, Yansong and Zheng, Xiawu and Ji, Jiayi and Zhang, Shengchuan and Cao, Liujuan},
  journal={arXiv preprint arXiv:2509.02560},
  year={2025}
}

@misc{litevggt,
      title={LiteVGGT: Boosting Vanilla VGGT via Geometry-aware Cached Token Merging}, 
      author={Zhijian Shu and Cheng Lin and Tao Xie and Wei Yin and Ben Li and Zhiyuan Pu and Weize Li and Yao Yao and Xun Cao and Xiaoyang Guo and Xiao-Xiao Long},
      year={2025},
      eprint={2512.04939},
      archivePrefix={arXiv},
      primaryClass={cs.CV},
      url={https://arxiv.org/abs/2512.04939}, 
}

@article{vggt-long,
  title={VGGT-Long: Chunk it, Loop it, Align it--Pushing VGGT's Limits on Kilometer-scale Long RGB Sequences},
  author={Deng, Kai and Ti, Zexin and Xu, Jiawei and Yang, Jian and Xie, Jin},
  journal={arXiv preprint arXiv:2507.16443},
  year={2025}
}

@misc{laser,
      title={LASER: Layer-wise Scale Alignment for Training-Free Streaming 4D Reconstruction}, 
      author={Tianye Ding and Yiming Xie and Yiqing Liang and Moitreya Chatterjee and Pedro Miraldo and Huaizu Jiang},
      year={2026},
      eprint={2512.13680},
      archivePrefix={arXiv},
      primaryClass={cs.CV},
      url={https://arxiv.org/abs/2512.13680}, 
}

@article{ttt3r,
  title={TTT3R: 3d reconstruction as test-time training},
  author={Chen, Xingyu and Chen, Yue and Xiu, Yuliang and Geiger, Andreas and Chen, Anpei},
  journal={arXiv preprint arXiv:2509.26645},
  year={2025}
}

@article{infinitevggt,
  title={InfiniteVGGT: Visual Geometry Grounded Transformer for Endless Streams},
  author={Yuan, Shuai and Yang, Yantai and Yang, Xiaotian and Zhang, Xupeng and Zhao, Zhonghao and Zhang, Lingming and Zhang, Zhipeng},
  journal={arXiv preprint arXiv:2601.02281},
  year={2026}
}

@article{dinov2,
  title={DINOv2: Learning Robust Visual Features without Supervision},
  author={Oquab, Maxime and Darcet, Timoth{\'e}e and Moutakanni, Th{\'e}o and Vo, Huy and Szafraniec, Marc and Khalidov, Vasil and Fernandez, Pierre and Haziza, Daniel and Massa, Francisco and El-Nouby, Alaaeldin and others},
  journal={{Transactions on Machine Learning Research Journal}},
  pages={1--31},
  year={2024}
}

@inproceedings{dpthead,
  title={Vision transformers for dense prediction},
  author={Ranftl, Ren{\'e} and Bochkovskiy, Alexey and Koltun, Vladlen},
  booktitle={{Proc. of CVPR}},
  pages={12179--12188},
  year={2021}
}

@inproceedings{tome,
  title={Token Merging: Your {ViT} but Faster},
  author={Bolya, Daniel and Fu, Cheng-Yang and Dai, Xiaoliang and Zhang, Peizhao and Feichtenhofer, Christoph and Hoffman, Judy},
  booktitle={ICLR},
  year={2023}
}

@misc{tomesd,
      title={Token Merging for Fast Stable Diffusion}, 
      author={Daniel Bolya and Judy Hoffman},
      year={2023},
      eprint={2303.17604},
      archivePrefix={arXiv},
      primaryClass={cs.CV},
      url={https://arxiv.org/abs/2303.17604}, 
}

@inproceedings{7scenes,
  title={Scene coordinate regression forests for camera relocalization in RGB-D images},
  author={Shotton, Jamie and Glocker, Ben and Zach, Christopher and Izadi, Shahram and Criminisi, Antonio and Fitzgibbon, Andrew},
  booktitle={{Proc. of CVPR}},
  pages={2930--2937},
  year={2013}
}

@inproceedings{nrgbd,
  title={Neural rgb-d surface reconstruction},
  author={Azinovi{\'c}, Dejan and Martin-Brualla, Ricardo and Goldman, Dan B and Nie{\ss}ner, Matthias and Thies, Justus},
  booktitle={{Proc. of CVPR}},
  pages={6290--6301},
  year={2022}
}

@inproceedings{scannet,
  title={Scannet: Richly-annotated 3d reconstructions of indoor scenes},
  author={Dai, Angela and Chang, Angel X and Savva, Manolis and Halber, Maciej and Funkhouser, Thomas and Nie{\ss}ner, Matthias},
  booktitle={{Proc. of CVPR}},
  pages={5828--5839},
  year={2017}
}

@inproceedings{tumdynamic,
  title={A benchmark for the evaluation of RGB-D SLAM systems},
  author={Sturm, J{\"u}rgen and Engelhard, Nikolas and Endres, Felix and Burgard, Wolfram and Cremers, Daniel},
  booktitle={Proc. of IROS},
  pages={573--580},
  year={2012},
  organization={IEEE}
}

@inproceedings{bonn,
  title={ReFusion: 3D reconstruction in dynamic environments for RGB-D cameras exploiting residuals},
  author={Palazzolo, Emanuele and Behley, Jens and Lottes, Philipp and Giguere, Philippe and Stachniss, Cyrill},
  booktitle={{Proc. of IROS}},
  pages={7855--7862},
  year={2019},
  organization={IEEE}
}

@article{kitti,
  title={Vision meets robotics: The kitti dataset},
  author={Geiger, Andreas and Lenz, Philip and Stiller, Christoph and Urtasun, Raquel},
  journal={{The International Journal of Robotics Research}},
  volume={32},
  number={11},
  pages={1231--1237},
  year={2013},
  publisher={Sage Publications Sage UK: London, England}
}

@inproceedings{sintel,
  title={A naturalistic open source movie for optical flow evaluation},
  author={Butler, Daniel J and Wulff, Jonas and Stanley, Garrett B and Black, Michael J},
  booktitle={Proc. of ECCV},
  pages={611--625},
  year={2012},
  organization={Springer}
}

@ARTICLE{lmm_3d_understanding,
  author={Wu, Yanmin and Gao, Qiankun and Zhang, Renrui and Li, Haijie and Zhang, Jian},
  journal={IEEE Transactions on Multimedia}, 
  title={Language-Assisted 3D Scene Understanding}, 
  year={2025},
  volume={27},
  number={},
  pages={3869-3879},
  doi={10.1109/TMM.2025.3535305}}

@ARTICLE{3ur_llm,
  author={Xiong, Haomiao and Zhuge, Yunzhi and Zhu, Jiawen and Zhang, Lu and Lu, Huchuan},
  journal={IEEE Transactions on Multimedia}, 
  title={3UR-LLM: An End-to-End Multimodal Large Language Model for 3D Scene Understanding}, 
  year={2025},
  volume={27},
  number={},
  pages={2899-2911},
  doi={10.1109/TMM.2025.3557620}}

@ARTICLE{virpnet,
  author={Wang, Lin and Sun, Shiliang and Zhao, Jing},
  journal={IEEE Transactions on Multimedia}, 
  title={VirPNet: A Multimodal Virtual Point Generation Network for 3D Object Detection}, 
  year={2024},
  volume={26},
  number={},
  pages={10597-10609},
  doi={10.1109/TMM.2024.3410117}}

@misc{streamvggt,
      title={Streaming 4D Visual Geometry Transformer}, 
      author={Dong Zhuo and Wenzhao Zheng and Jiahe Guo and Yuqi Wu and Jie Zhou and Jiwen Lu},
      year={2026},
      eprint={2507.11539},
      archivePrefix={arXiv},
      primaryClass={cs.CV},
      url={https://arxiv.org/abs/2507.11539}, 
}

@inproceedings{gao2024hicom,
  title = {{HiCoM}: Hierarchical Coherent Motion for Dynamic Streamable Scenes with {3D} Gaussian Splatting},
  author={Gao, Qiankun  and Meng, Jiarui and Wen, Chengxiang  and Chen, Jie and Zhang, Jian},
  booktitle = {Proc. of NeurIPS},
  year = {2024}
}

@inproceedings{fu2025recon,
   title={ReCon-GS: Continuum-Preserved Gaussian Streaming for Fast and Compact Reconstruction of Dynamic Scenes}, 
   author={Jiaye Fu and Qiankun Gao and Chengxiang Wen and Yanmin Wu and Siwei Ma and Jiaqi Zhang and Jian Zhang},
   booktitle = {Proc. of NeurIPS},
   year = {2025}
}

@inproceedings{lu2024scaffold,
  title={Scaffold-gs: Structured 3d gaussians for view-adaptive rendering},
  author={Lu, Tao and Yu, Mulin and Xu, Linning and Xiangli, Yuanbo and Wang, Limin and Lin, Dahua and Dai, Bo},
  booktitle={Proc. of CVPR},
  pages={20654--20664},
  year={2024}
}

@article{yang2026robo3r,
  title={Robo3R: Enhancing Robotic Manipulation with Accurate Feed-Forward 3D Reconstruction},
  author={Yang, Sizhe and Xu, Linning and Li, Hao and Mu, Juncheng and Zeng, Jia and Lin, Dahua and Pang, Jiangmiao},
  journal={arXiv preprint arXiv:2602.10101},
  year={2026}
}

@inproceedings{vggt-omega,
  title={VGGT-{$\Omega$}},
  author={Wang, Jianyuan and Chen, Minghao and Zhang, Shangzhan and Karaev, Nikita and Sch{\"o}nberger, Johannes and Labatut, Patrick and Bojanowski, Piotr and Novotny, David and Vedaldi, Andrea and Rupprecht, Christian},
  booktitle={Proc. of CVPR},
  year={2026}
}

\begin{IEEEbiographynophoto}
{Zecheng Tang} is currently pursuing the M.S. degree in computer science and technology with Peking University Shenzhen Graduate School, Shenzhen, China. His research interests include: 1) 3D reconstruction  and neural rendering; 2) 3D vision-language multimodal learning.
\end{IEEEbiographynophoto}

\begin{IEEEbiographynophoto}
    {Jiaye Fu} is currently pursuing the M.S. degree in computer science and technology with Peking University Shenzhen Graduate School, Shenzhen, China. His research interests include image/video coding, 3D reconstruction, and AIGC.
\end{IEEEbiographynophoto}

\begin{IEEEbiographynophoto}
    {Qiankun Gao} is currently a Postdoctoral Researcher with the School of Electronic and Computer Engineering, Peking University, Shenzhen, China. His research interests include 3D vision, AI-generated content (AIGC), and continual learning.
\end{IEEEbiographynophoto}

\begin{IEEEbiographynophoto}
    {Haijie Li} is currently a Master's student at Peking University, advised by Prof. Jian Zhang, and he will start his Ph.D. program in September 2026. His research interests include: 1) 3D reconstruction and neural rendering; 2) 3D vision-language multimodal learning.
\end{IEEEbiographynophoto}

\begin{IEEEbiographynophoto}
    {Yanmin Wu} received the Ph.D. degree in computer applied technology from the School of Electronic and Computer Engineering, Peking University, Shenzhen, China, in 2025. His research interests include 3D vision and vision-language multimodal learning, autonomous driving, and robotics.
\end{IEEEbiographynophoto}

\begin{IEEEbiographynophoto}
    {Jiaqi Zhang} is now a Associate Researcher in the School of Computer Science, Peking University. His research interests include data compression and image/video coding.
\end{IEEEbiographynophoto}

\begin{IEEEbiographynophoto}
    {Siwei Ma} (Fellow, IEEE) joined the School of Computer Science, Peking University, where he is currently a Professor. He has authored more than 300 technical articles in refereed journals and proceedings in image and video coding, video processing, video streaming, and transmission. He served/serves as an Associate Editor for the IEEE Transactions on Circuits and Systems for Video Technology and Journal of Visual Communication and Image Representation.
\end{IEEEbiographynophoto}

\begin{IEEEbiographynophoto}
    {Jian Zhang} (Member, IEEE) is an Associate Professor and leads the Visual-Information Intelligent Learning LAB (VILLA) at the School of Electronic and Computer Engineering, Peking University, Shenzhen, China. He has published over 120 technical articles in refereed international journals and proceedings and has received over 16000 citations. He serves as an Associate Editor for the Journal of Visual Communication and Image Representation.
\end{IEEEbiographynophoto}

\vfill

\end{document}